\documentclass[10pt]{asme2ej}
\usepackage{graphicx} 
\usepackage{epsfig}
\usepackage{fancyhdr}
\usepackage{setspace}
\usepackage{helvet}
\usepackage[hyphens]{url}
\usepackage{amsmath,bm}
\usepackage{gensymb}
\usepackage{calc}
\usepackage{amssymb}
\usepackage{array}
\usepackage{arydshln} 

\usepackage{caption}
\usepackage{subcaption}
\usepackage{lipsum}
\usepackage{footnote}
\usepackage{mathrsfs}
\usepackage{algorithm}
\usepackage{algpseudocode}
\usepackage{balance}
\usepackage{xcolor}
\usepackage{soul}
\usepackage{multirow}

\algnewcommand\algorithmicinput{\textbf{Variables:}}
\algnewcommand\Variables{\item[\algorithmicinput]}

\DeclareCaptionFont{blue}{\color{blue}}

\renewcommand{\footnoterule}{%
    \kern -3pt
    \hrule width \textwidth height 0.5pt
    \kern 2pt%
}

\usepackage[square,numbers]{natbib}

\title{Kinetostatic Optimization for Kinematic Redundancy Planning of Nimbl'Bot Robot }

\author{Angelica Ginnante$^{1,2,3}$, Stéphane Caro$^1$, Enrico Simetti$^{2}$,
\affiliation{
{\tensfb François Leborne$^3$}\\
$^1$Nantes Université, École Centrale Nantes, CNRS, LS2N, UMR 6004, F-44000 Nantes, France\\
$^2$University of Genova, DIBRIS, 16145 Genova, Italy\\
$^3$Nimbl'Bot, 7 Avenue de Guitayne, 33610 Canéjan, France\\
Email: aginnante@nimbl-bot.com, Stephane.Caro@ls2n.fr, Enrico.Simetti@unige.it,\\ fleborne@nimbl-bot.com
}
}

\begin{document}

\maketitle
\setlength{\parindent}{1ex}
\begin{abstract}
\textit{\it
In manufacturing industry, Computer Numerical Control~(CNC) machines are often preferred over Industrial Serial Robots~(ISR) for machining tasks.
Indeed, CNC machines offer high positioning accuracy, which leads to slight dimensional deviation on the final product.
However, these machines have a restricted workspace generating limitations in the machining work.
Conversely, ISR are typically characterized by a larger workspace.
ISR have already shown satisfactory performance in tasks like polishing, grinding and deburring.
This paper proposes a kinematic redundant robot composed of a novel two degrees-of-freedom mechanism with a closed kinematic chain.
After describing a task priority inverse kinematic control framework used for joint trajectory planning exploiting the robot kinematic redundancy, the paper analyses the kinetostatic performance of this robot depending on the considered control tasks.
Moreover, two kinetostatic tasks are introduced and employed to improve the robot performance.
Simulation results show how the robot better performs when the optimization tasks are active.
}
\end{abstract}

\section{Introduction}

Machining tasks in manufacturing industry were automated over the past few decades using automated machining tools to speed up processes, such as milling and grinding, and improve their quality~\cite{liang2004machining}.
Machining tasks are often performed by Computer Numerical Control~(CNC) machines, as such industrial machines feature a very good accuracy.
Nevertheless, these machines are generally expensive and do not provide a high versatility~\cite{Ji2019industrial}.
Therefore, Industrial Serial Robots~(ISR) started to be investigated for manufacturing tasks since they can cover larger workspaces and increase the range of achievable operations, but their real capabilities in machining applications have yet to be realized as explained in~\cite{chen2013robot}.
Furthermore, ISR reduces the scrap rates and production costs compared to CNC machines~\cite{caro2013workpiece}.
ISR have already shown satisfactory performance in some tasks, like polishing~\cite{Takeuchi1993automated}, grinding~\cite{Liu1990robotic} and deburring~\cite{Pires2002force}.
The topic of ISR in machining tasks is actively studied and new techniques have been defined to improve their performance, for example, analysing their stiffness~\cite{dumas2011joint,dumas2012joint}.
Moreover, the authors of~\cite{brunete2018hard} collected a series of tools to define a solid background for building ISR employed in industrial machining applications.

In machining trajectory planning, the kinematic redundancy of robotic manipulators can be viewed as a possible way to improve the machine abilities and performance~\cite{gonul2019improved}.
In fact, the kinematic redundancy allows working in cluttered environments such as medical robotics~\cite{menon2017trajectory}.
Moreover, the kinematic redundancy can be used for solving several tasks simultaneously while optimizing some performance criteria~\cite{seereeram1995global}.
From a mathematical point of view, a robot is kinematically redundant with respect to a task when it has more degrees-of-freedom than the required amount necessary to perform that task~\cite{Siciliano1990kinematic}.
Namely, the desired task can be achieved by an infinite number of possible robot configurations.
Redundant robots can be designed in several ways and are classified into three categories: discrete, continuous and modular robots~\cite{chirikjian1992geometric,chirikjian1995kinematically}.
In~\cite{Shammas2003new}, the authors presented a novel two degrees-of-freedom mechanism for modular redundant robots.
The investigated concept uses a complex design optimized for compactness, strength and range of motion.
However, the proposed mechanism was never used to build a machining robot.
This paper introduces a new patented two degrees-of-freedom mechanism employed as actuator for redundant manipulators designed by Nimbl'Bot~\cite{patent} and named NB-module.

The main drawback of redundant robots is solving the inverse kinematic problem~\cite{Chirikjian1992hyper_redundant}.
The simplest way to deal with this problem is using the pseudo-inverse Jacobian matrix method~\cite{Khalil2004modeling}.
However, this approach neither avoids singularities nor takes advantage of the robot kinematic redundancy to perform robot configuration optimization.
In~\cite{reiter2016inverse}, the authors treated the problem of trajectory planning for a general kinematic redundant robot as two interdependent problems, inverse kinematics and trajectory optimization, to identify the time-optimal path tracking solution.
However, this method does not solve other simultaneous tasks, e.g., performance criteria improvement or obstacle avoidance.
The work presented in~\cite{dharmawan2018task} describes the solution for kinematically redundant robots given a set of desired tasks and performance metrics through the use of a mixed analytical and numerical approaches.
This method is compared with other types, like genetic algorithm~\cite{marcos2010evolutionary} and neural networks~\cite{toshani2014real}, demonstrating an improvement in the time computation.
However, this methodology does not present any hierarchy associated to the tasks or mechanisms for activating and deactivating the tasks.
This reduces the amount of possible tasks that can be taken into account by the control algorithm.
In~\cite{escande2014hierarchical}, the authors present a hierarchical quadratic programming control algorithm used to find a solution to multiple and antagonistic objectives for humanoid robot motion generation.
Another multiple tasks control framework is presented in~\cite{di2019handling}, called Set-Based Multi-Task Priority Inverse Kinematics Framework.
In~\cite{flacco2012prioritized}, the so-called Saturation in the Null Space~(SNS) algorithm implements a predictive prioritizing technique for multiple tasks.
One of the main drawbacks of these task priority algorithms is that activating or deactivating one or more tasks generates a discontinuity in the null space projector~\cite{Simetti2016inequality}.

This paper introduces the Nimbl'Bot two degrees-of-freedom mechanism and presents its geometric and kinematic models.
The complete computation of these models is shown in~\cite{ginnante2021design}.
Then, the first design of the ``Nimbl'Bot robot" composed of a serial arrangement of ten mechanisms is described and tested by tracing several trajectories to analyze its kinetostatic performance.
To kinematically control this robot, a task-priority based control algorithm, named Task Priority Inverse Kinematic (TPIK)~\cite{Simetti2016inequality,Simetti2018priority,Simetti2019priority}, is employed.
Moreover, two new tasks, based on kinetostatic performance metrics, are added to the control algorithm to improve the robot performance.
The paper aims to demonstrate how a kinematically redundant robot can be employed in machining task trajectory planning while optimizing its kinetostatic performance.

The paper is organized as follows, Section~\ref{SEC::nimb_rob} presents the Nimbl'Bot robot design and shows the first prototype.
Section~\ref{SEC::nimb_mech} introduces the Nimbl'Bot actuation mechanism, its geometric and kinematic models and its workspace.
Section~\ref{SEC::tpik} describes the kinematic control algorithm used to test the robot.
Section~\ref{SEC::kinetostatic_metrics} defines the metrics used to evaluate the robot kinetostatic performance and the Jacobian matrices that relate the rate of these metrics with the robot joint velocities.
Section~\ref{SEC::test} presents the trajectory planning of the Nimbl'Bot robot for a series of trajectories and compares the obtained kinetostatic results.
Conclusions and future work are given in Section~\ref{SEC::conc}.

\section{Nimbl'Bot Robot Overview}
\label{SEC::nimb_rob}
\begin{figure}[!b]
    \centering
    \includegraphics[scale=0.5]{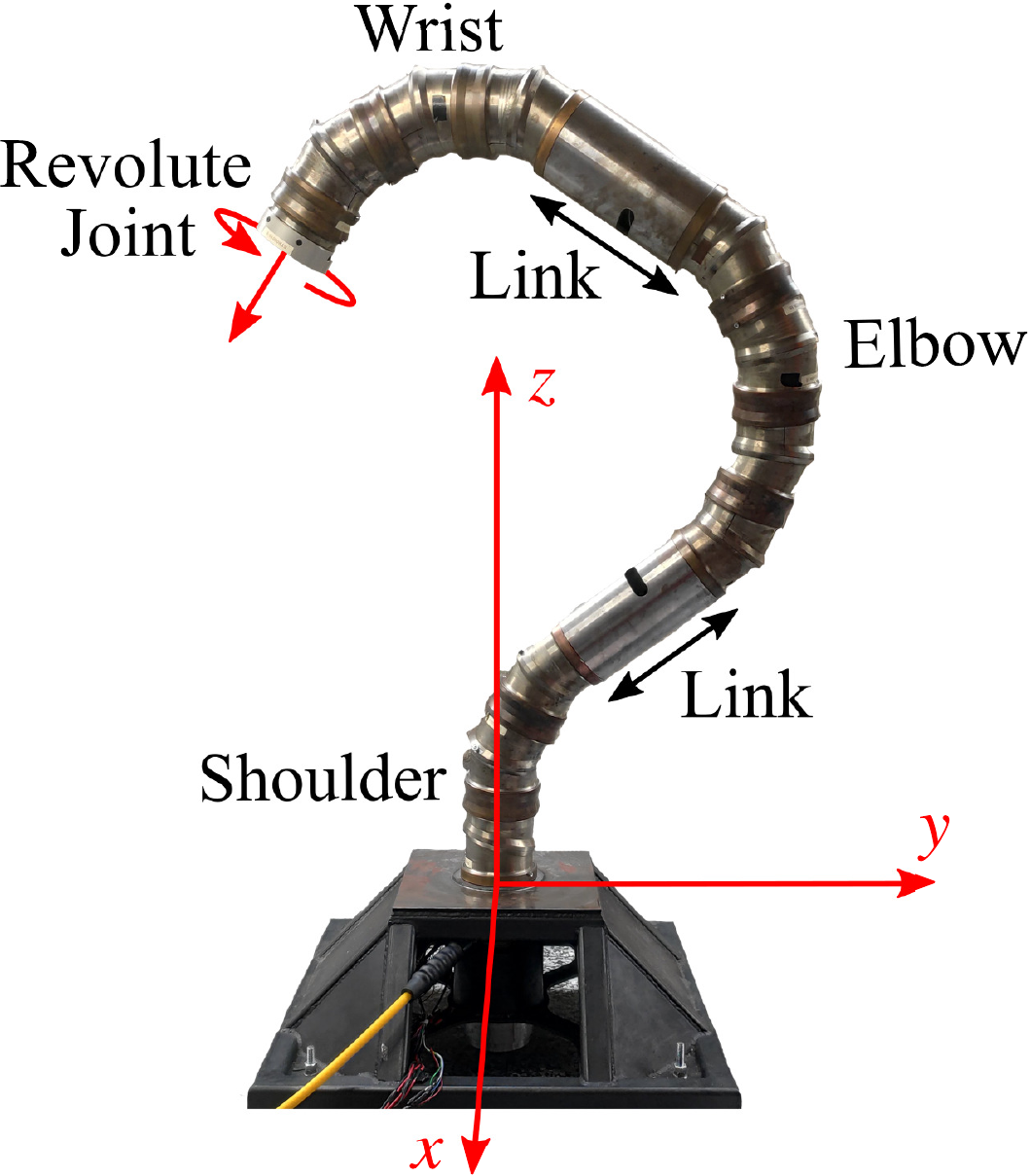}
    \caption{RP-120 actuated by ten NB-modules mounted in series and a final revolute joint. Shoulder and wrist made of three NB-modules, total covered solid angle of~$\pm \pi/2$ radians each. Elbow made of four NB-modules, solid angle of~$\pm 2\pi/3$ radians.}
    \label{FIG::c_r}
\end{figure}

The first prototype of the Nimbl'Bot robot developed for machining is shown in Fig.~\ref{FIG::c_r}.
This robot is composed by the serial arrangement of ten NB-modules.
Each NB-module base platform is attached to the previous NB-module ending platform.
This version of the Nimbl'Bot robot is called RP-120.
It was design for machining application, for example, the milling, grooving and trimming of small metal parts for the nuclear sector.

\begin{figure}[!t]
    \centering
    \includegraphics[scale=0.65]{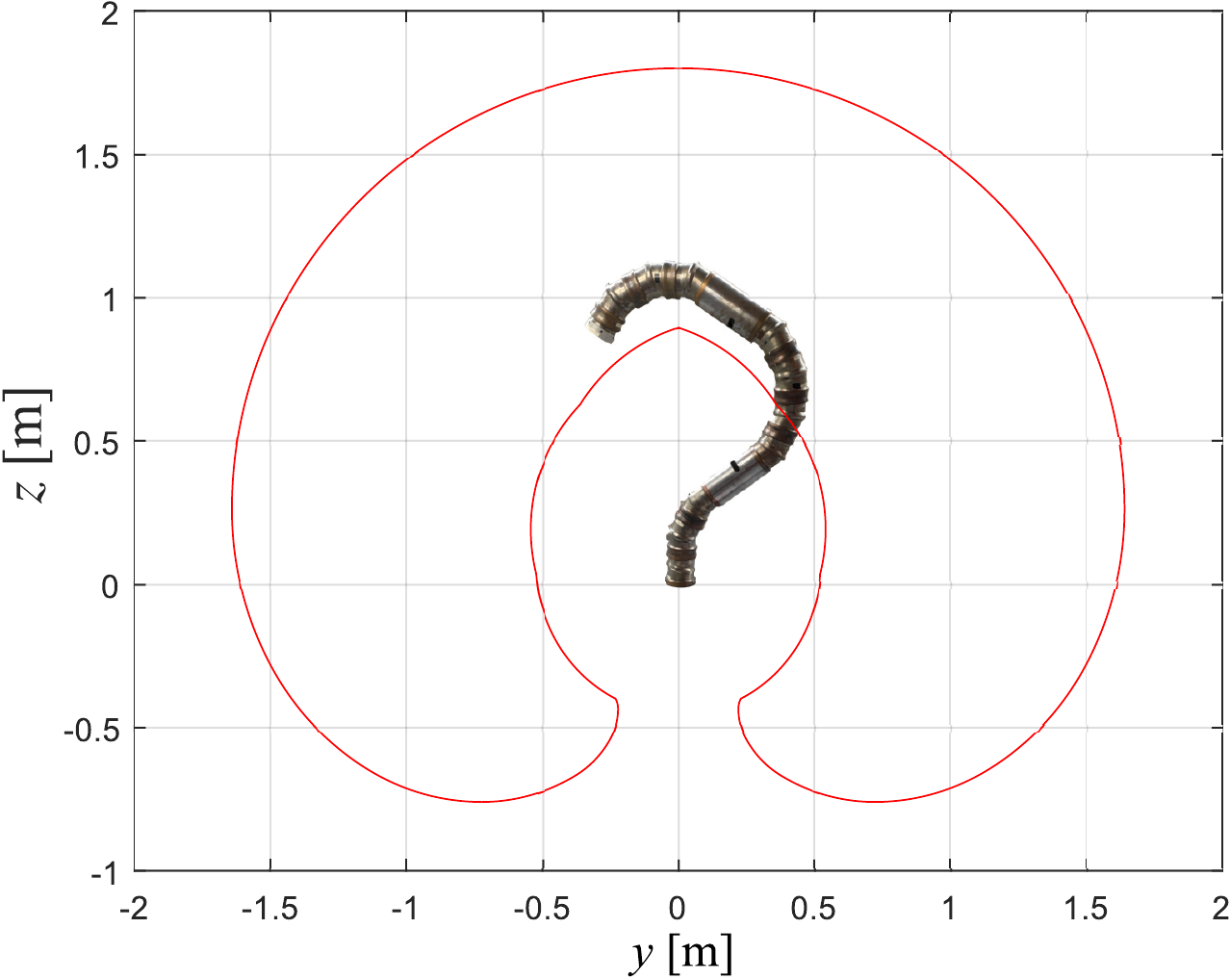}
    \caption{Vertical section of the RP-120 Cartesian workspace for at least one end-effector orientation}
    \label{FIG::rob_workspace}
\end{figure}

The RP-120 can be divided into three regions, i.e. the shoulder, composed of three NB-modules, the elbow, composed of four NB-modules, and the wrist, composed of three NB-modules.
Since the solid angle reachable by one NB-module is~$\pm \pi/6$, several NB-modules need to be arranged together to ensure a sufficiently large orientation workspace.
Figure~\ref{FIG::rob_workspace} shows the vertical section of the RP-120 Cartesian workspace reachable by at least one end-effector orientation.
The complete workspace is obtained by rotating this section around the~$z$~axis.
In total, the RP-120 has 21~degrees-of-freedom since each of the ten NB-modules provides two degrees-of-freedom, plus a final revolute joint that allows adjusting the tool orientation.
So, this design is redundant considering that the task of following a trajectory, typical of machining operations, only requires five degrees-of-freedom.
Mathematically, a robot is defined as redundant when the dimension of its actuation vector~$\mathbf{q}\in\mathbb{R}^n$ is greater than the dimension of the task vector~$\mathbf{x}\in\mathbb{R}^m$, namely when~$n>m$~\cite{chirikjian1994hyper}.

\begin{figure}[!t]
\centering
\begin{subfigure}[b]{.2\textwidth}
    \centering
    \includegraphics[scale=0.21]{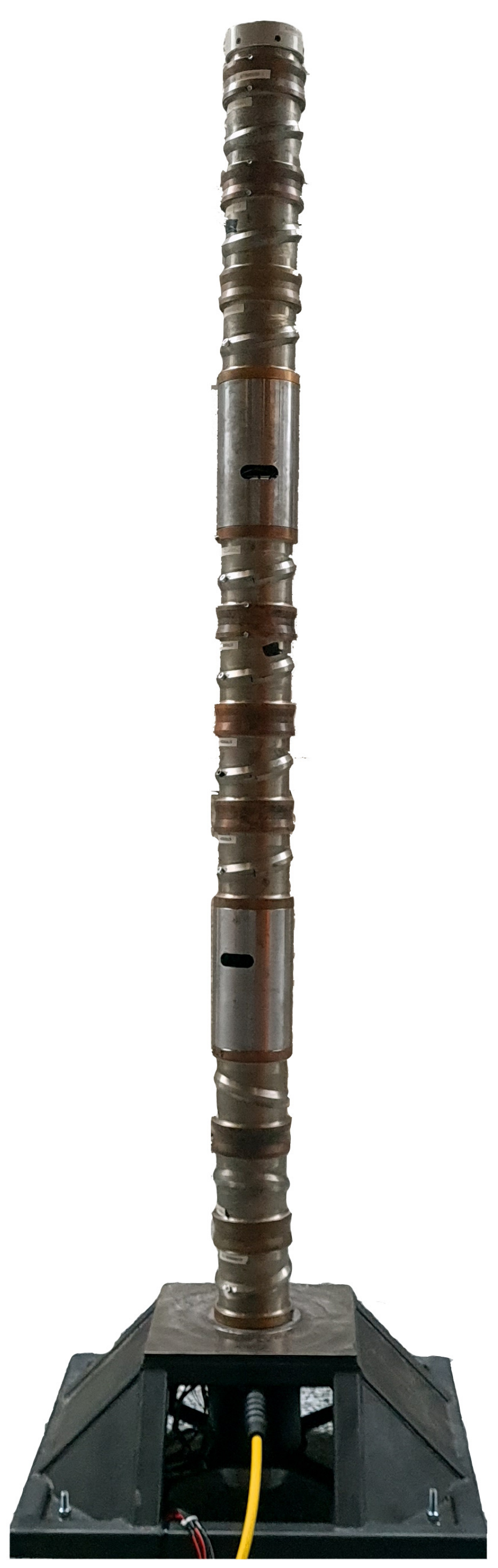}
    \caption{}
    \label{FIG::real_robot_straight}
\end{subfigure}%
\hfill
\begin{subfigure}[b]{.35\textwidth}
    \centering
    \includegraphics[scale=0.45]{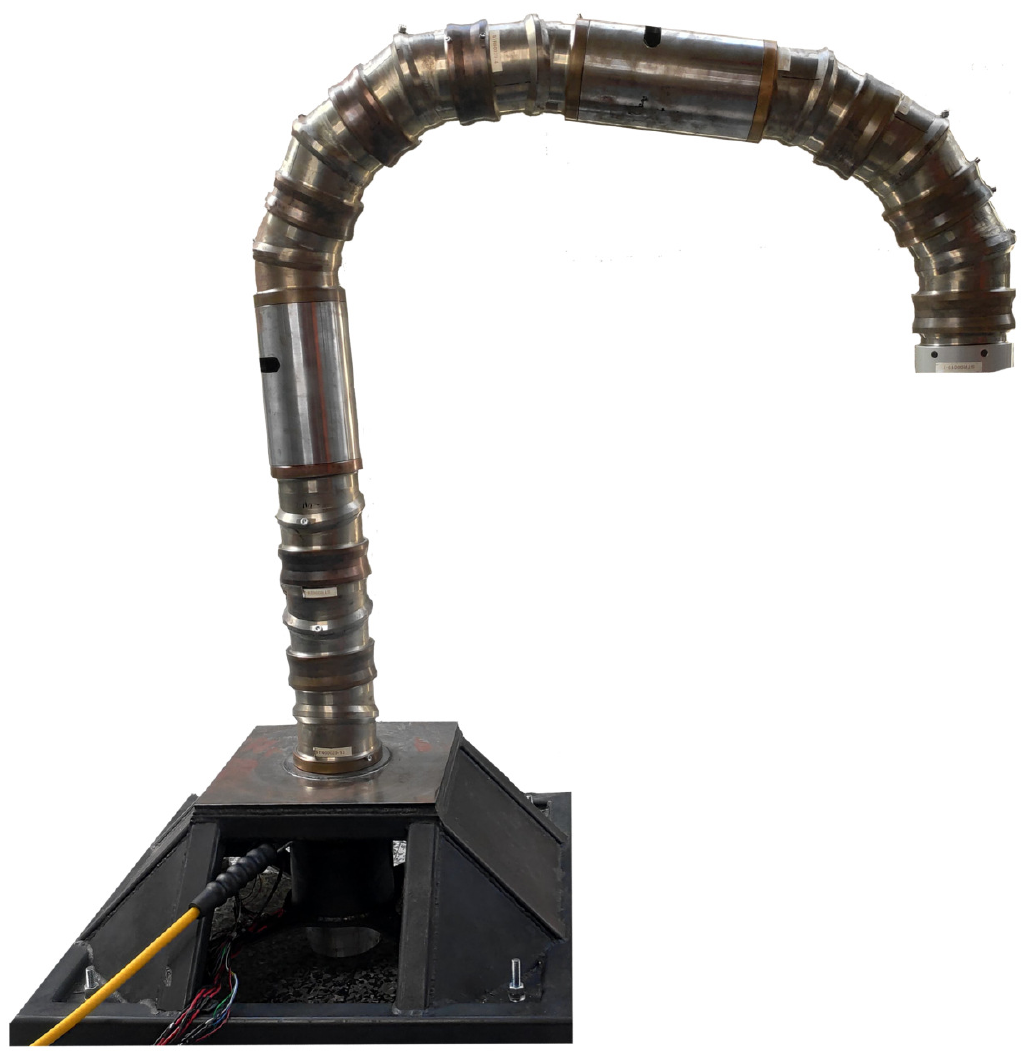}
    \caption{}
    \label{FIG::real_robot_bended1}
\end{subfigure}%
\hfill
\begin{subfigure}[b]{.2\textwidth}
    \centering
    \includegraphics[scale=0.44]{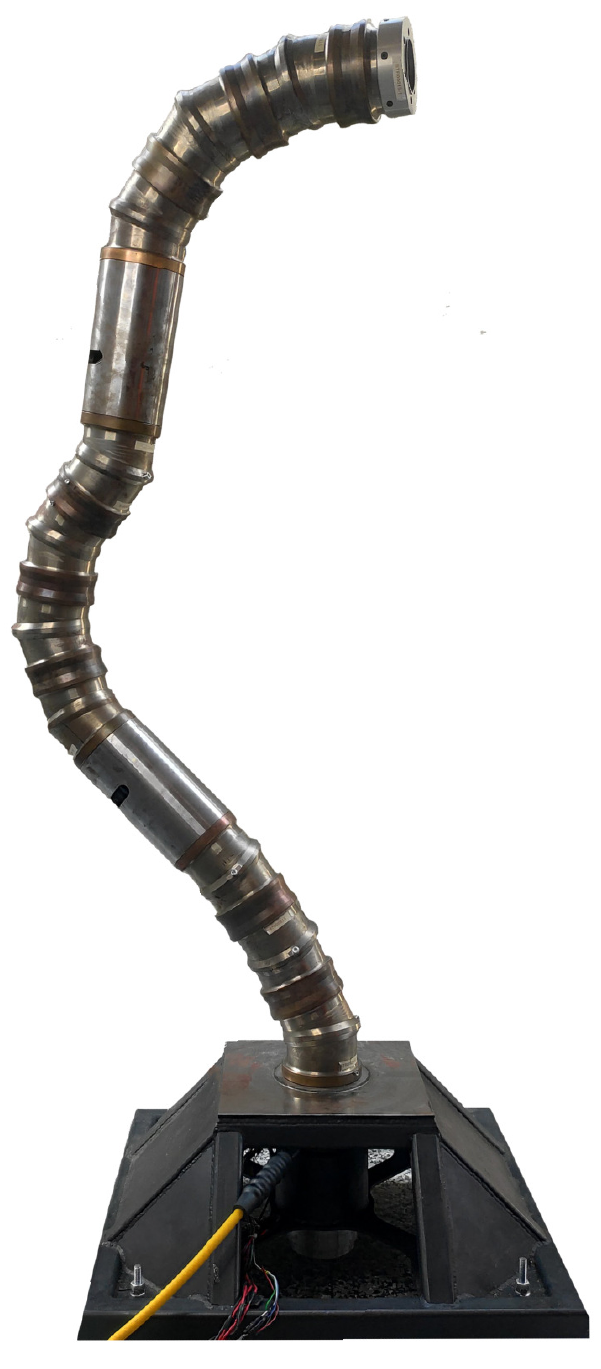}
    \caption{}
    \label{FIG::real_robot_bended2}
\end{subfigure}%
\hfill
\begin{subfigure}[b]{.25\textwidth}
    \centering
    \includegraphics[scale=0.46]{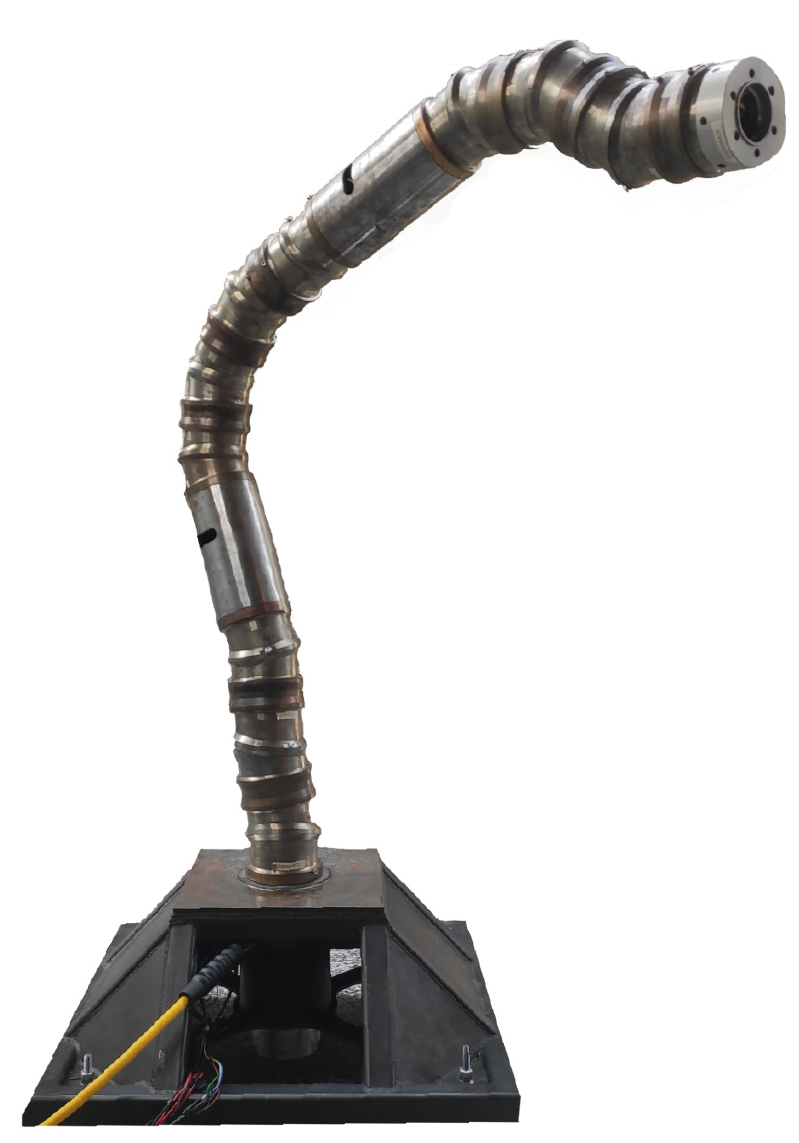}
    \caption{}
    \label{FIG::real_robot_bended3}
\end{subfigure}
\caption{Four postures of the RP-120 prototype}
\label{FIG::real_robot}
\end{figure}

Figure~\ref{FIG::real_robot} shows four postures of the RP-120 first prototype.
A cable passes inside the RP-120 and constrains the architecture cancelling most of the mechanical backlash.
The two links are hollow to reduce the weight and allow the routing of internal cables.

\section{Nimbl'Bot Module Overview}
\label{SEC::nimb_mech}

This section describes the novel two degrees-of-freedom mechanism developed by the company Nimbl'Bot and called NB-module.
This actuation system features a closed kinematic chain composed of two chains, one internal and the other external.
It is actuated by two motors that provide two rotational motions.
The NB-module geometric and kinematic models are analysed in~\cite{ginnante2021design}.

\subsection{Description of the External Kinematic Chain}

\begin{figure}[!t]
\centering
\begin{minipage}[b]{.49\textwidth}
    \centering
    \includegraphics[scale=0.18]{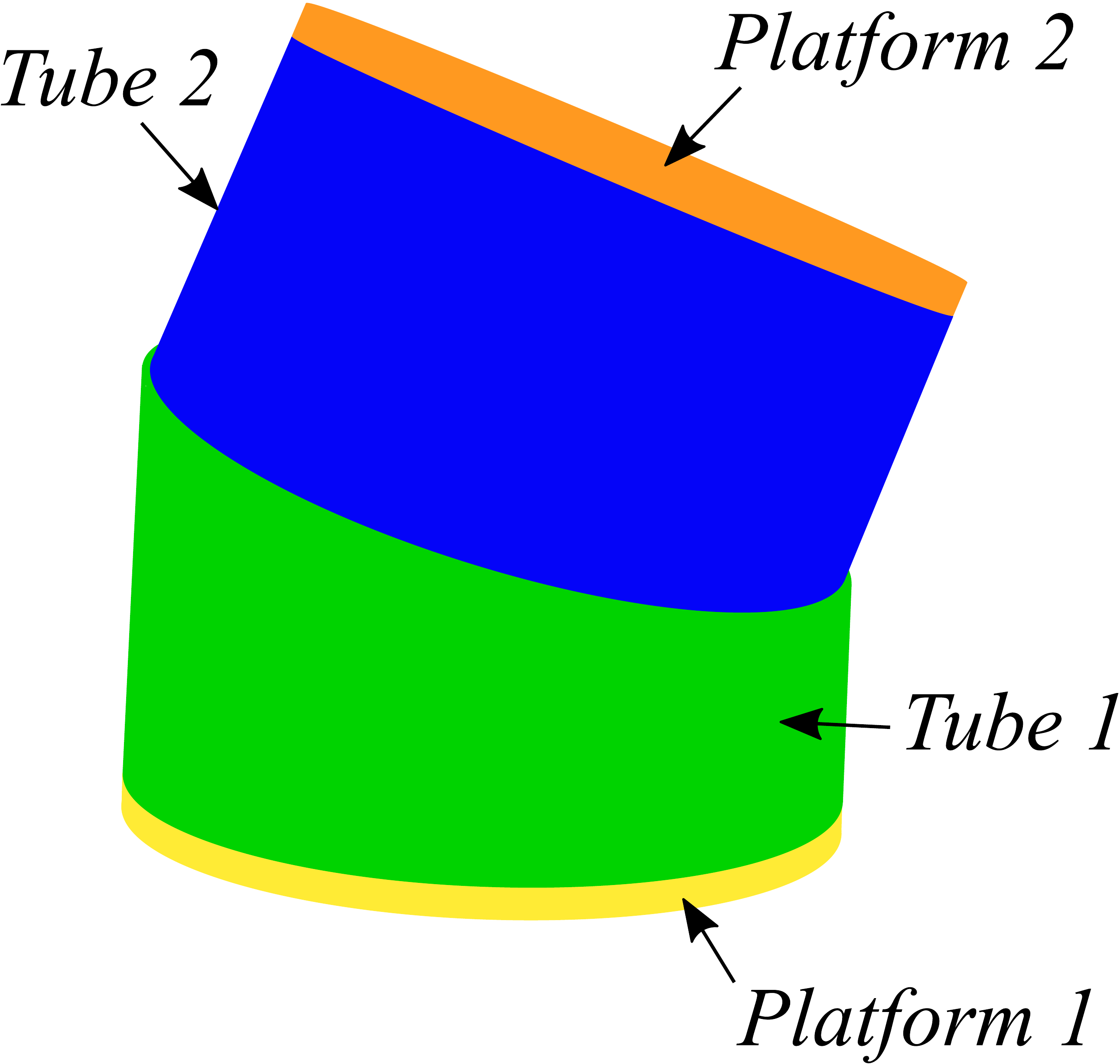}
    \caption{External view of the NB-module}
    \label{FIG::ext}
\end{minipage}%
\begin{minipage}[b]{.49\textwidth}
    \centering
    \includegraphics[scale=0.165]{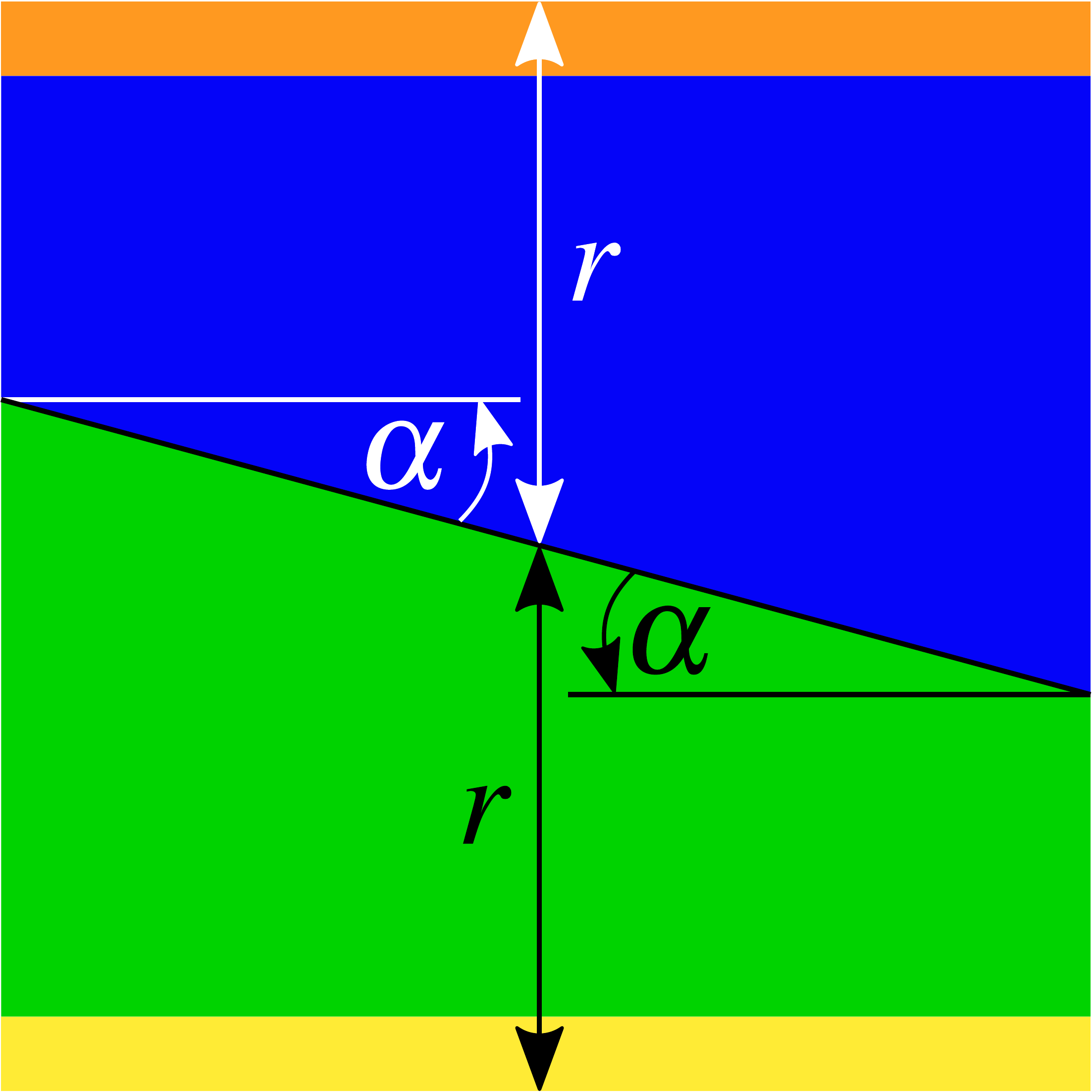}
    \caption{Front view of the external kinematic chain}
    \label{FIG::ext_front}
\end{minipage}%
\end{figure}

The external kinematic chain has seven different components.
Four of them are shown in Fig.~\ref{FIG::ext}.
The fixed base, in yellow, is named \textit{Platform~1} and is considered centered on the origin frame for which the NB-module transformation matrix is calculated.
Above \textit{Platform~1}, there is a rotating cylinder, in green, named \textit{Tube~1}.
\textit{Tube~1} is a hollow cylinder cut by an oblique plane with height~$r$ and slope~$\alpha$, shown in Fig.~\ref{FIG::ext_front}.
The first motor actuates the component \textit{Tube~1}.
The motor is attached directly to the inner side of \textit{Tube~1}.
In this way, \textit{Tube~1} can rotate around the vertical axis that passes through the center of \textit{Platform~1}.
\textit{Tube~2} is placed over \textit{Tube~1} and they have the same shape in this case, even if, in principle, their height~$r$ and slope~$\alpha$ could be different.
The second motor actuates \textit{Tube~2} and is attached to the inner side of \textit{Tube~2}.
\textit{Tube~2} can rotate around the axis perpendicular to and centered in \textit{Platform~2}.
The external kinematic chain is closed by \textit{Platform~2}, the moving platform, which is the end-effector of the NB-module.

\begin{figure}[!t]
\centering
\begin{minipage}[b]{.49\textwidth}
    \centering
    \includegraphics[scale=0.23]{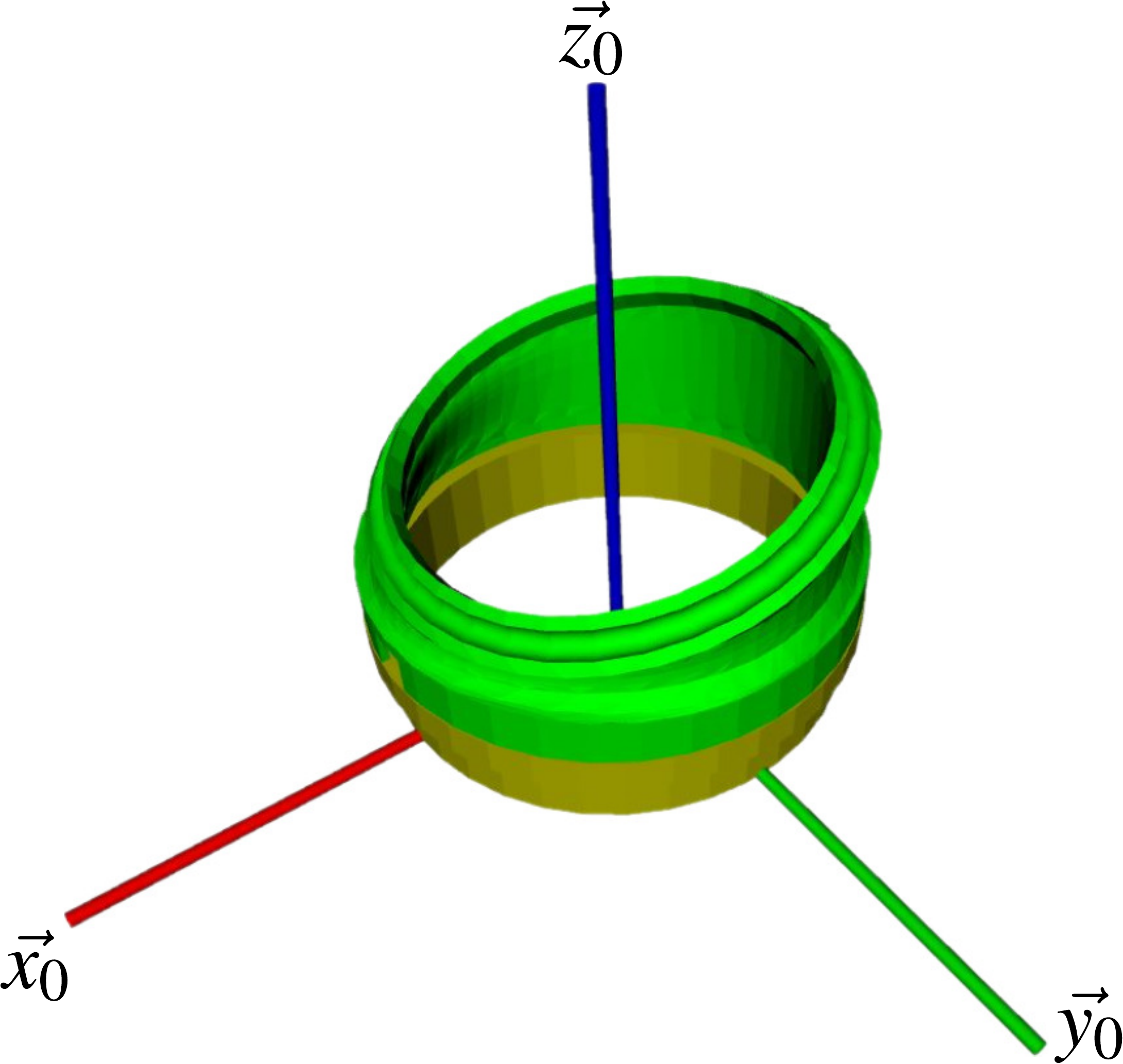}
    \caption{View of \textit{Tube~1} oblique side}
    \label{FIG::circle_tube1}
\end{minipage}%
\begin{minipage}[b]{.49\textwidth}
    \centering
    \includegraphics[scale=0.23]{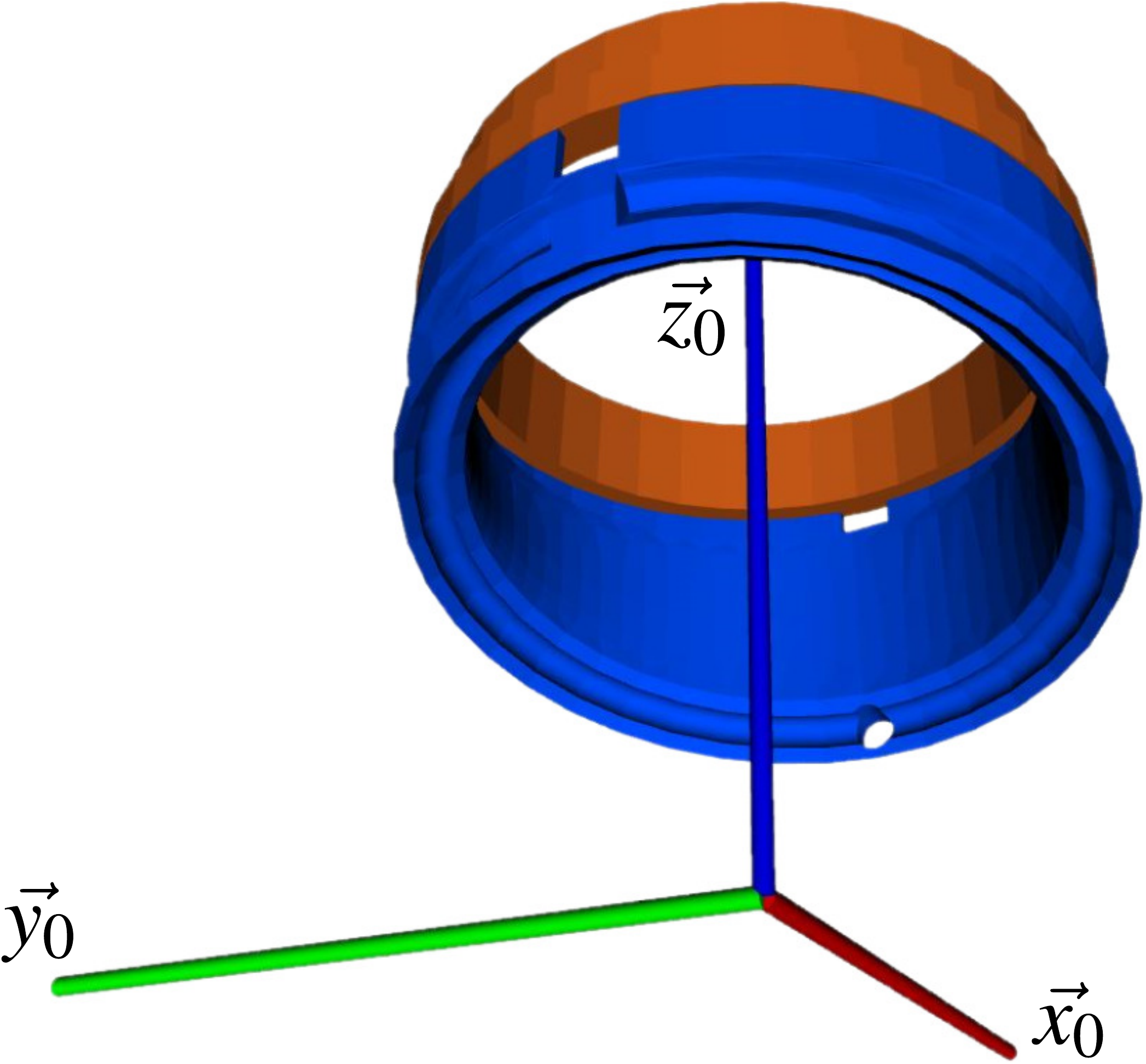}
    \caption{View of \textit{Tube~2} oblique side}
    \label{FIG::circle_tube2}
\end{minipage}
\end{figure}

Cutting obliquely the cylindrical tubes results in an elliptical shape.
However, the tube oblique sides are reshaped in a circular way to allow a continuous rotation between the oblique planes.
So, as it can be seen in Figs.~\ref{FIG::circle_tube1} and~\ref{FIG::circle_tube2}, a circular groove is designed above the inclined sides of the two tubes.

\begin{figure}[!t]
     \centering
     \begin{subfigure}[b]{0.3\textwidth}
         \centering
    	 \includegraphics[scale=0.23]{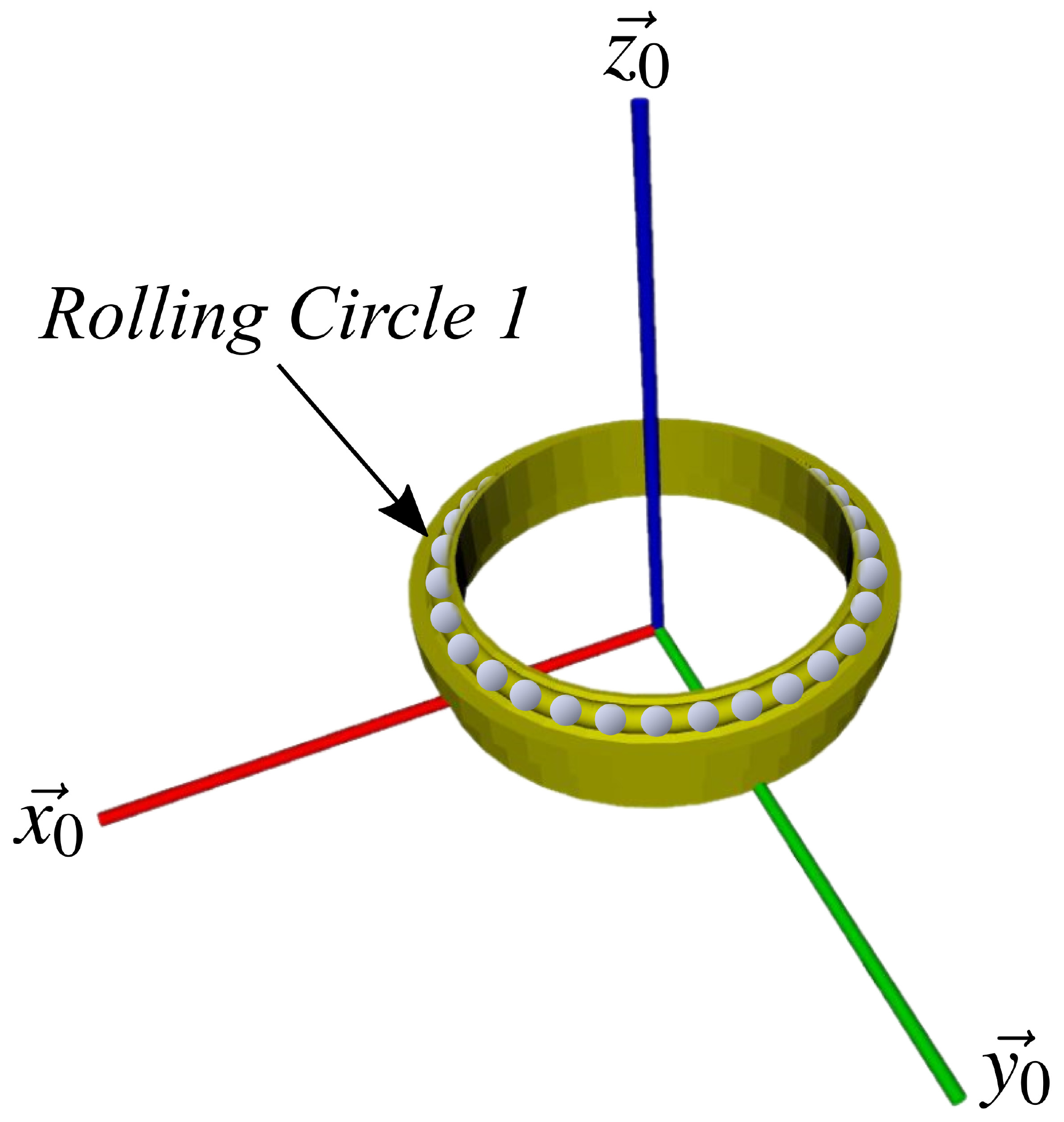}
         \caption{Location of \textit{Rolling~Circle~1} over \textit{Platform~1}}
         \label{FIG::P1_Balls}
     \end{subfigure}
     \hfill
     \begin{subfigure}[b]{0.3\textwidth}
         \centering
    	 \includegraphics[scale=0.23]{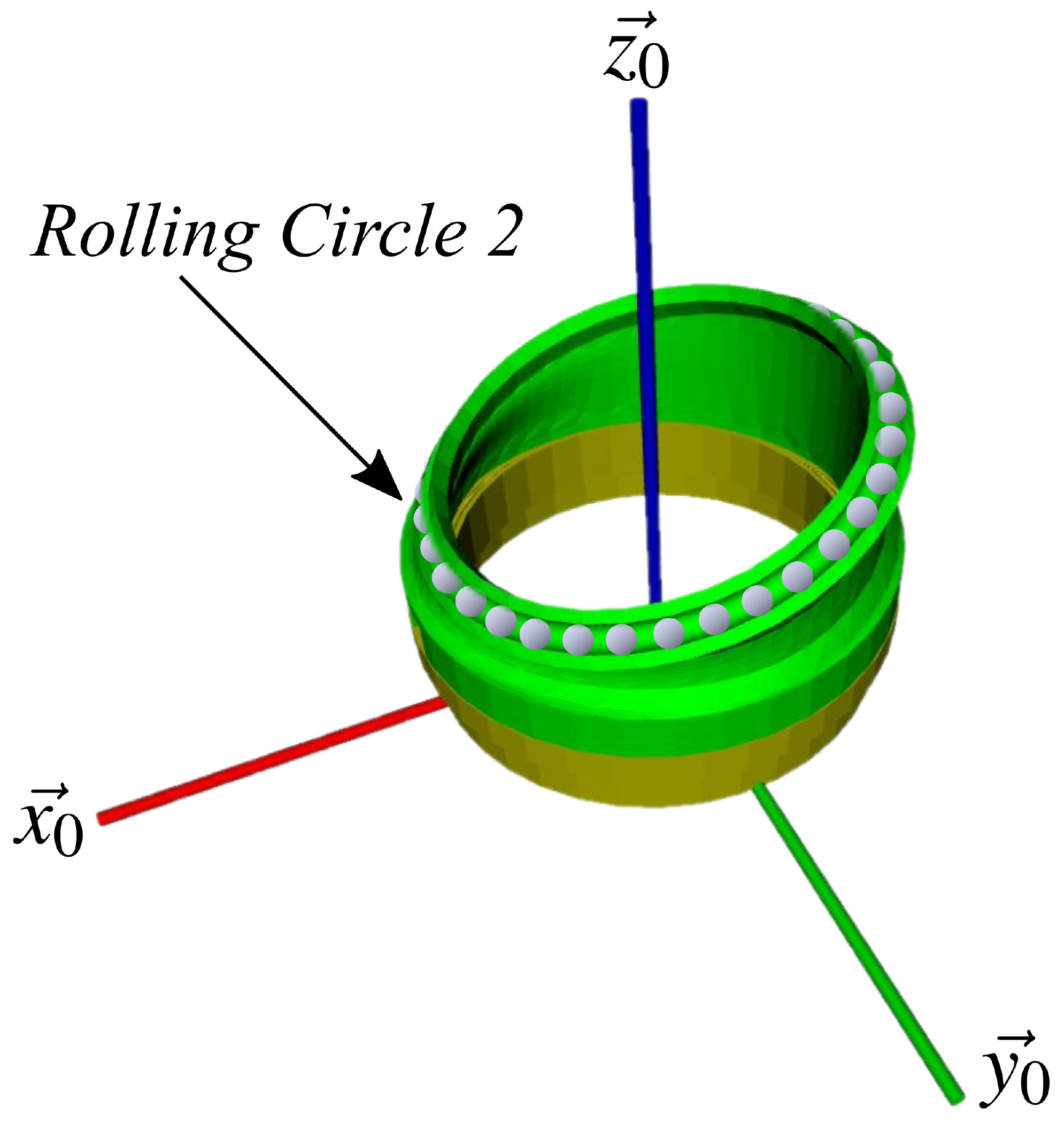}
         \caption{Location of \textit{Rolling~Circle~2} over \textit{Tube~1}}
         \label{FIG::T1_Balls}
     \end{subfigure}
     \hfill
     \begin{subfigure}[b]{0.3\textwidth}
    	 \centering
    	 \includegraphics[scale=0.23]{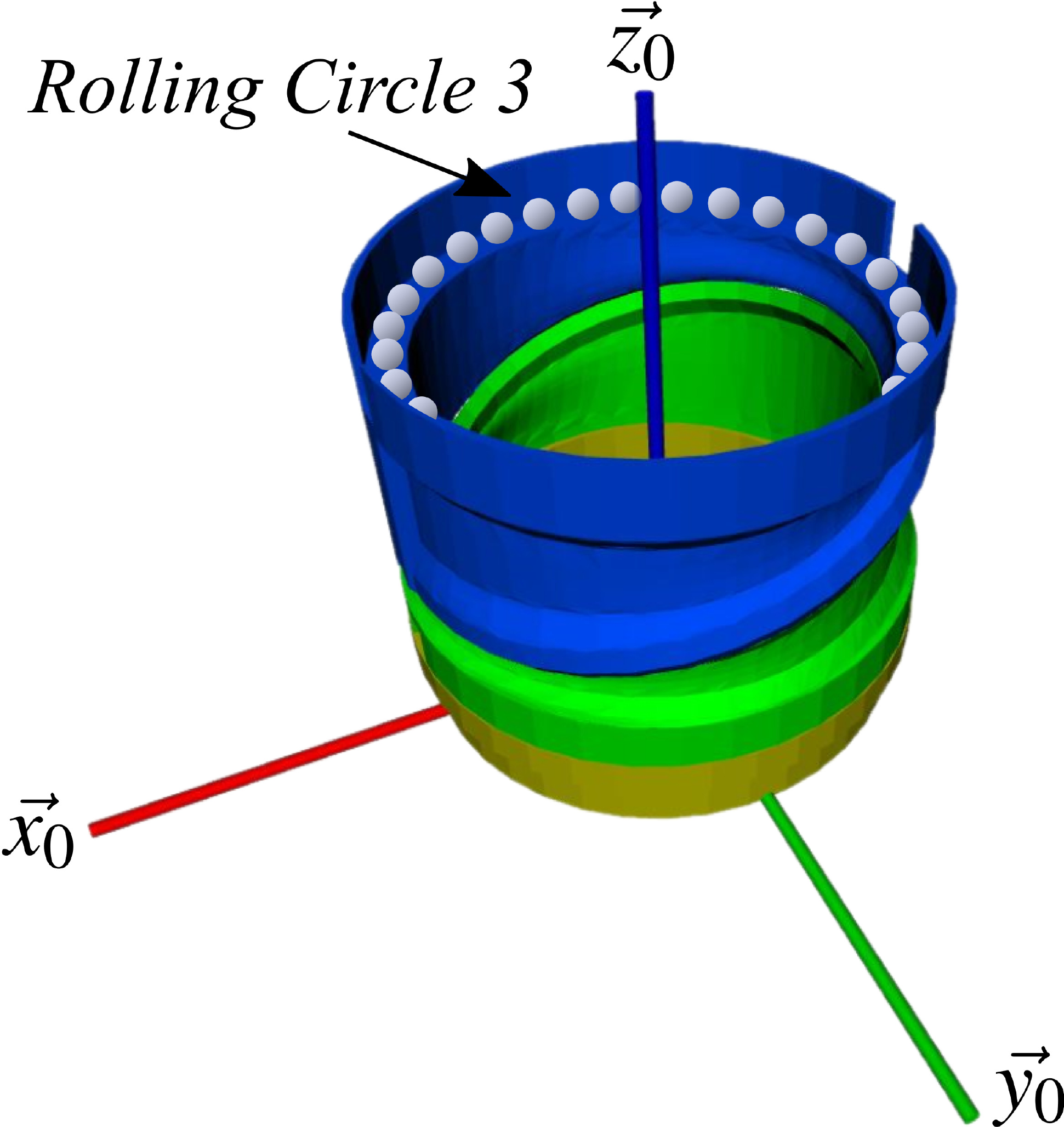}
         \caption{Location of \textit{Rolling~Circle~3} over \textit{Tube~2}}
         \label{FIG::T2_Balls}
     \end{subfigure}
     \hfill
     \caption{Location of rolling circles formed of a series of balls in the NB-module}
     \label{FIG::Balls}
\end{figure}

Three rolling circles formed of a series of small balls are inserted between the platforms and the tubes to allow a fluid movement.
The balls are inserted in the grooves machined in the platforms and tubes, as shown in Fig.~\ref{FIG::Balls}.
These are called \textit{Rolling~Circle~1}, \textit{Rolling~Circle~2} and \textit{Rolling~Circle~3} and represent the last three elements of the external kinematic chain.
Consequently, \textit{Tube~1} and \textit{Tube~2} can independently rotate with a continuous movement.

\subsection{Description of the Internal Kinematic Chain}

\begin{figure}[!t]
    \centering
    \includegraphics[scale=0.2]{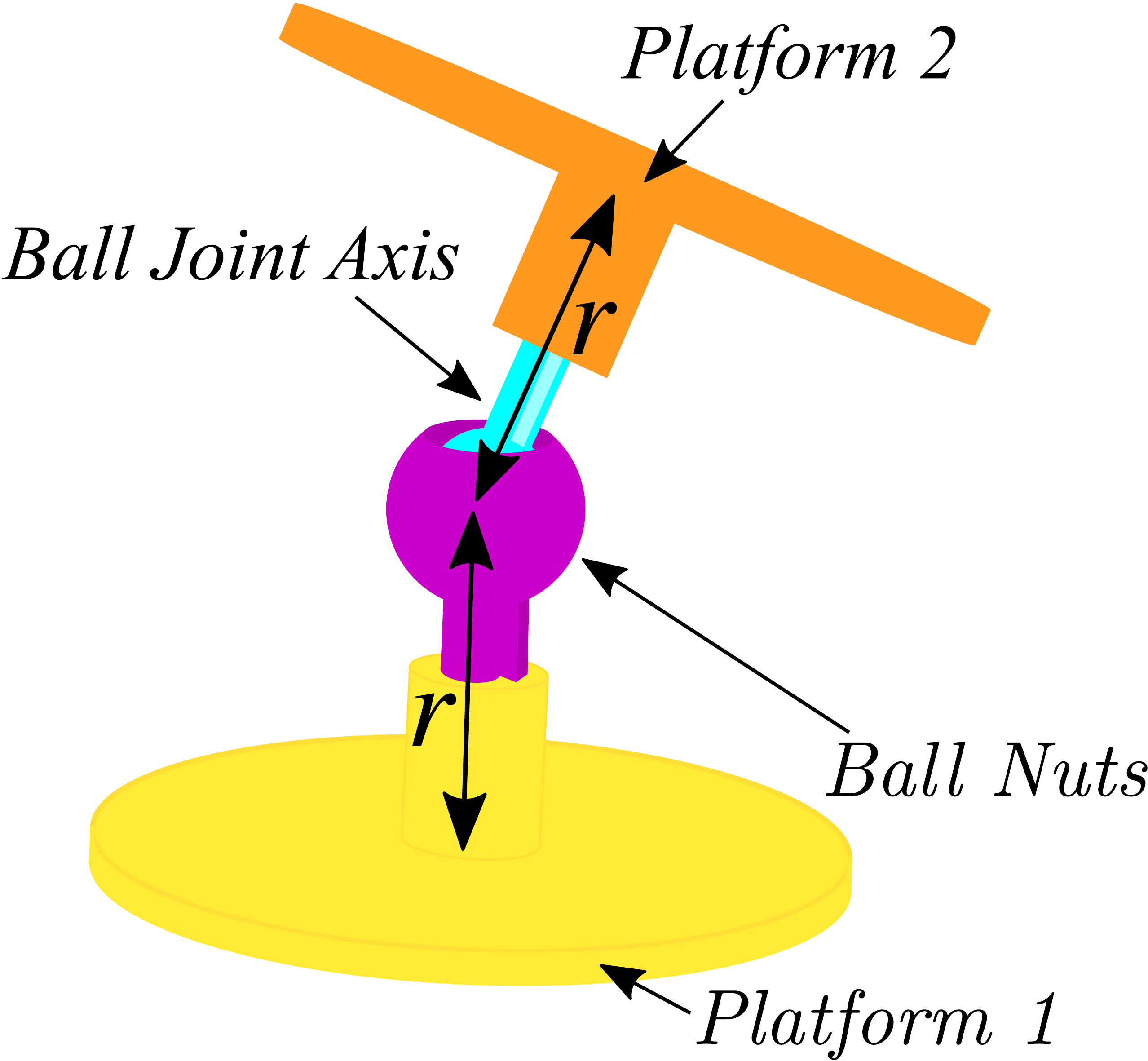}
    \caption{Internal view of the NB-module}
    \label{FIG::int}
\end{figure}

The internal kinematic chain has four components, as shown in Fig.~\ref{FIG::int}.
Two of those also belong to the external chain, i.e. \textit{Platform~1} and \textit{Platform~2}, generating the closed kinematic chains mechanism.
\textit{Platform~1} is linked to the component \textit{Ball~Nut} in purple through a prismatic joint, which prevents internal breaks while the NB-module is actuated.
These could occur due to dimensional inaccuracies in the mechanical parts.
Following that, there is \textit{Ball~Joint~Axis} in cyan, which forms a constant velocity joint with \textit{Ball~Nut}.
Finally, the \textit{Ball~Joint~Axis} is linked to \textit{Platform~2} through another prismatic joint, again to avoid internal breaks.
The variable~$r$ equal to the tube heights represents also the distance between \textit{Platform~1} and the constant velocity joint and between the constant velocity joint and \textit{Platform~2}.

The interesting feature of the NB-module is the constant velocity joint~\cite{Carricato2009homokinetic}, which works like a universal joint, but allows its two ends to rotate at the same velocity.
Therefore, fixing \textit{Platform~1} means forcing no torsion to the whole internal kinematic chain.
So, thanks to the constant velocity joint and the rolling circles that decouple the rotation of each component, the tube rotations lead to an inclination of \textit{Platform~2} with no rotation about its normal axis.
The NB-module amounts to a zero-torsion mechanism.
 
\subsection{Geometric Model}

\begin{figure}[!t]
    \centering
    \includegraphics[scale=0.3]{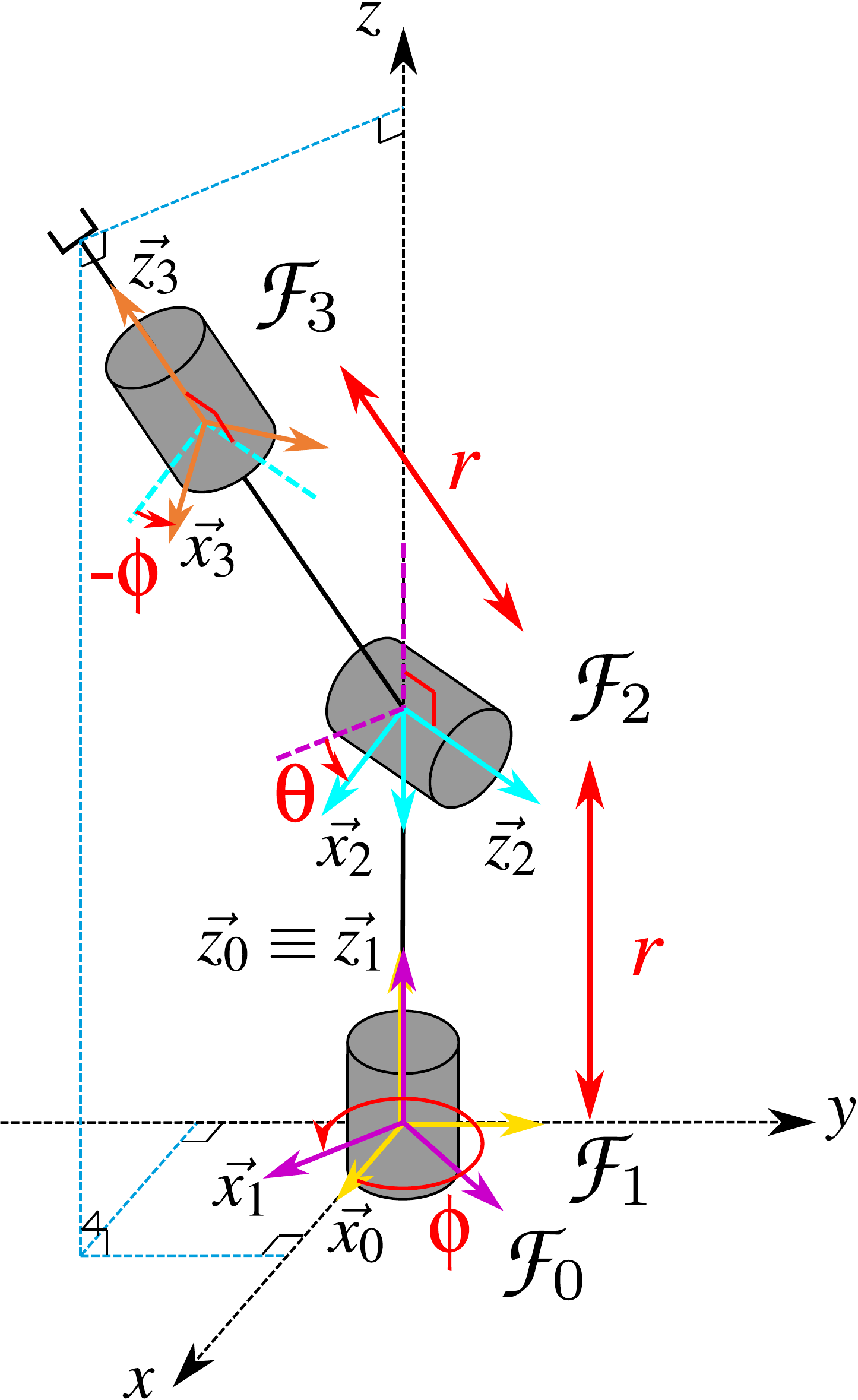}
    \caption{Tilt and azimuth model of the NB-module}
    \label{FIG::scheme_t&a}
\end{figure}

\begin{figure}[!t]
    \centering
\begin{minipage}{.49\textwidth}
    \centering
    \includegraphics[scale=0.23]{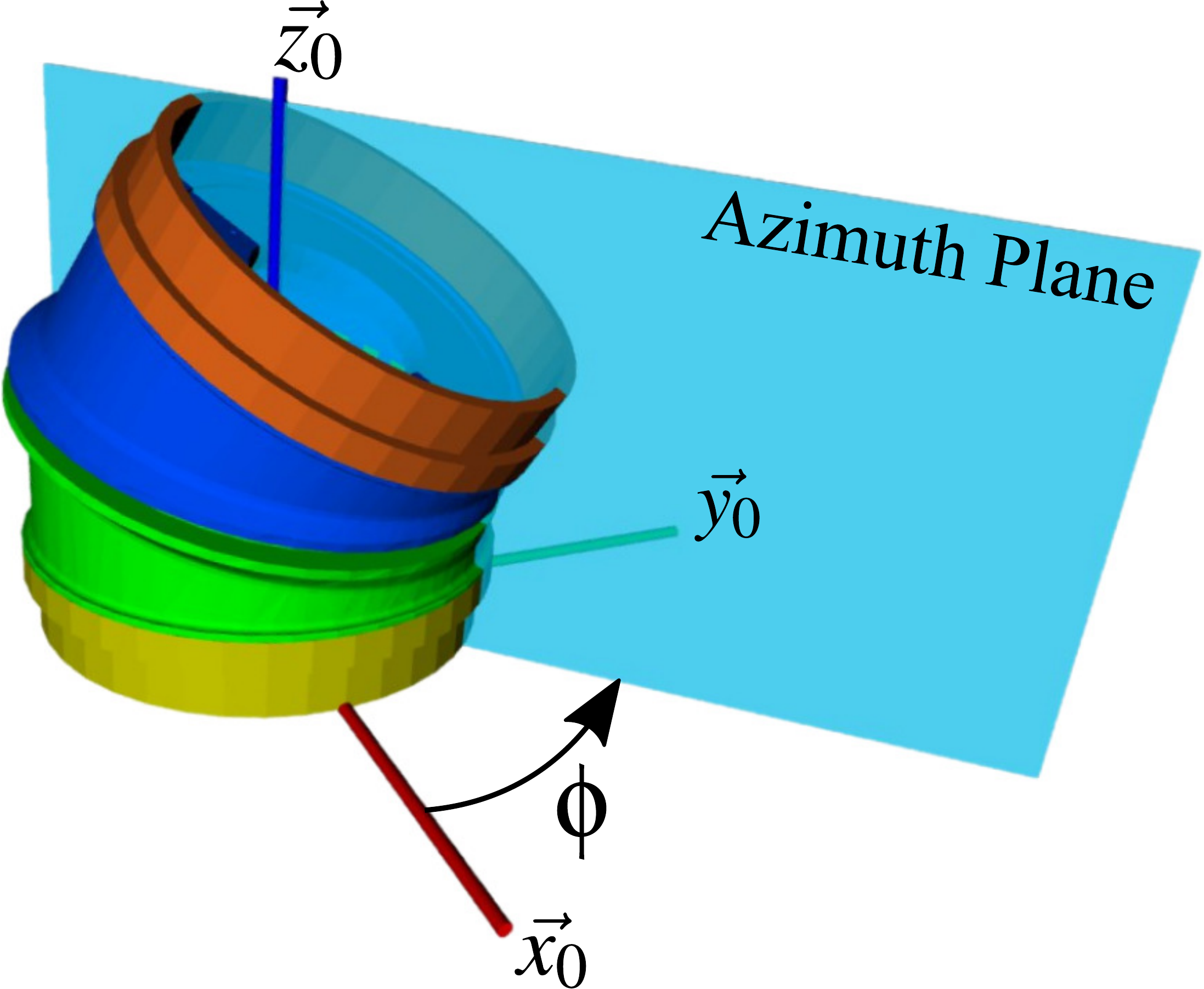}
    \caption{Azimuth plane}
    \label{FIG::a_planes}
\end{minipage}%
\begin{minipage}{.49\textwidth}
    \centering
    \includegraphics[scale=0.23]{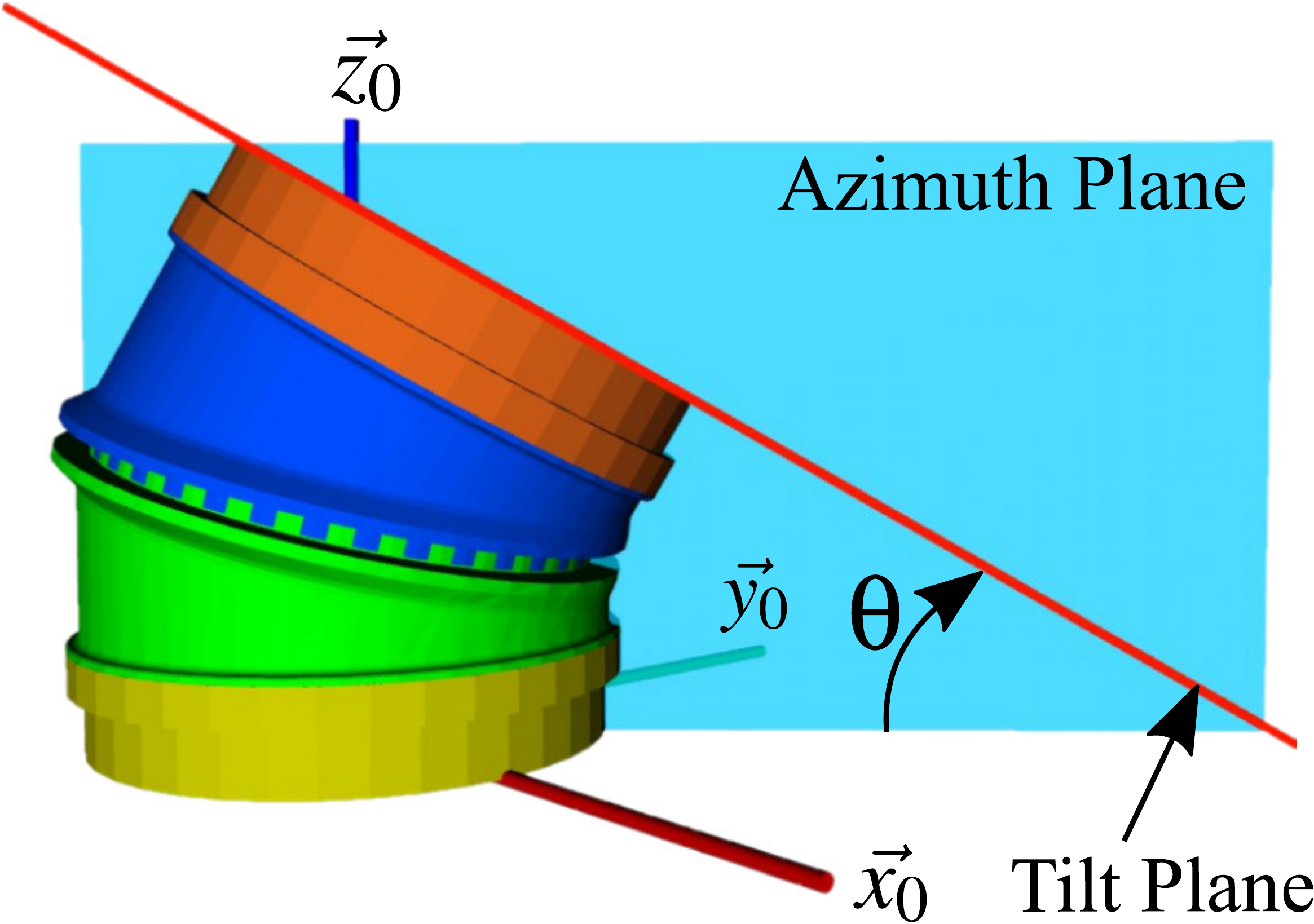}
    \caption{Tilt and azimuth planes}
    \label{FIG::t_a_planes}
\end{minipage}%
\end{figure}

Here, the geometric model of the NB-module is described.
Since the NB-module is a zero-torsion mechanism, the rotation matrix of a frame rigidly attached to \textit{Platform~2} with respect to a frame rigidly attached to \textit{Platform~1} can be described using the notation presented in~\cite{Bonev2002advantages}, i.e. the Tilt \& Azimuth (T\&A) angle notation.
Figure~\ref{FIG::scheme_t&a} shows the T\&A based geometric model of the NB-module.
A series of three revolute joints form the model.
The first revolute joint represents the azimuth angle~$\phi$ of the NB-module.
The second revolute joint is the tilt angle~$\theta$.
The third revolute joint is constrained to have the negative value of the azimuth angle~$-\phi$ as a consequence of the NB-module zero-torsion characteristics.
The azimuth angle~$\phi$ gives the orientation along which \textit{Platform~2} is tilted.
Figure~\ref{FIG::a_planes} shows the Azimuth Plane.
Figure~\ref{FIG::t_a_planes} shows the Tilt Plane, which is parallel to the top of the \textit{Platform~2} and oriented along with the Azimuth Planes.
The angle between the plane spanned by axes~$\vec{x_0}$ and~$\vec{y_0}$ and the Tilt Plane is the tilt angle~$\theta$.

Given the kinematic chain of Fig.~\ref{FIG::scheme_t&a}, the NB-module rotation matrix~$^0\mathbf{R}_3(\phi, \theta)$ and translation vector~$^0\mathbf{p}_3(\phi, \theta, r)$ pointing from the origin of frame~$\mathcal{F}_0$ to the origin of frame~$\mathcal{F}_3$ are
\begin{equation}
    \mathbf{R}(\phi,\theta)=\begin{bmatrix}
        \cos^2{\phi} \cos{\theta} + \sin^2{\phi}\;&&
        \cos{\phi} \sin{\phi} (\cos{\theta} - 1)\;&&
        \cos{\phi} \sin{\theta}\\
        \sin{\phi} \cos{\phi} (\cos{\theta} - 1)\;&&
        \sin^2{\phi} \cos{\theta} + \cos^2{\phi}\;&&
        \sin{\phi} \sin{\theta}\\
        -\sin{\theta} \cos{\phi}\;&&
        -\sin{\theta} \sin{\phi}\;&&
	    \cos{\theta}\\
    \end{bmatrix}
    \:\:\:\text{and}\:\:\:
    ^0\mathbf{p}_3(\phi, \theta, r) =
    \left[
    \begin{array}{c}
    r \sin{\theta} \cos{\phi} \\
    r \sin{\theta} \sin{\phi} \\
    r + r \cos{\theta} \\
    \end{array}
    \right].
\label{EQ::spherical}
\end{equation}

The complete homogeneous transformation matrix of the NB-module from frame~$\mathcal{F}_0$ to frame~$\mathcal{F}_3$ is expressed as
\begin{equation}
    ^0\mathbf{T}_3(\phi, \theta, r) =
    \left[
        \begin{array}{ccc|c}
            &  &  &  \\
            & ^0\mathbf{R}_3(\phi, \theta) &  & ^0\mathbf{p}_3(\phi, \theta, r) \\
            & &  &  \\
            \hline
            & \mathbf{0}_{3 \times 1} & & 1
        \end{array}
    \right].
\label{EQ::transf_mat_TA}
\end{equation}

The azimuth~$\phi$ and tilt~$\theta$ angles help express the NB-module transformation matrix in the T\&A notation.
However, these angles do not represent the angular position of the motors attached to the inner side of each tube.
The angular positions of the motors are called~$q_1$ and~$q_2$
So, the azimuth~$\phi$ and tilt~$\theta$ rotation angles need to be expressed as functions of the actuation variables~$q_1$ and~$q_2$.
Similarly,~$q_1$ and~$q_2$ can be expressed as functions of~$\phi$ and~$\theta$.
\begin{equation}
    \begin{cases}
        \phi = \displaystyle\frac{q_1+q_2-\pi}{2}\\
        \theta = \arctan\left(-\displaystyle\frac{2\tan{\alpha} \sin\left(\frac{q_1-q_2}{2}\right)}{1-\tan^2{\alpha} \sin^2\left(\frac{q_1-q_2}{2}\right)}\right)
    \end{cases}\quad\quad
    \begin{cases}
        q_1 = \phi+ \arccos\left(-\displaystyle\frac{\cos{\alpha}\left(\cos{\theta}-1\right)}{\sin{\alpha} \sin{\theta}}\right) \\
        q_2 =  \phi- \arccos\left(-\displaystyle\frac{\cos{\alpha}\left(\cos{\theta}-1\right)}{\sin{\alpha} \sin{\theta}}\right)+\pi
    \end{cases},
     \label{EQ::a&t_1j}
\end{equation}
where~$\alpha$ is the slope of the oblique planes in \textit{Tube~1} and \textit{Tube~2}.
When the tilt~$\theta$ is equal to 0, the value of the actuation variables is~$q_1 $~$=$~$ q_2 $~$=$~$ \phi + \pi / 2$.

\subsection{Kinematic Model}

This section presents the kinematic model of the NB-module based on the parameterization defined in Fig.~\ref{FIG::scheme_t&a}.
The kinematic Jacobian matrix of the NB-module is computed in two steps.
First of all, the Jacobian matrix~${\bf J}_1(\phi, \theta, r)\in\mathbb{R}^{6 \times 2}$ is calculated as a function of the angles~$[\phi,\theta]^\top$.
This matrix maps the tilt and azimuth angles time derivatives~$\left[\dot{\phi}, \dot{\theta} \right]^\top$ to the NB-module tip twist~$\mathbf{t}$~$=$~$\left[\dot{\mathbf{p}}^\top,\bm{\omega}^\top\right]^\top\in \mathbb{R}^6$ where~$\dot{\mathbf{p}}\in \mathbb{R}^3$ and~$\bm{\omega}\in \mathbb{R}^3$ are the linear and angular velocity vectors of frame~$\mathcal{F}_3$, respectively.
The Jacobian~$\mathbf{J}_1$ results to be
\begin{equation}
    \mathbf{t} = {\bf J}_1(\phi, \theta, r)
    \begin{bmatrix}
        \dot{\phi} & \dot{\theta}
    \end{bmatrix}^\top\:\:\:\text{with}\:\:\:
    {\bf J}_1(\phi, \theta, r) = \left[
    \begin{array}{cc}
    - r \sin{\phi} \sin{\theta} & r \cos{\phi} \cos{\theta} \\
      r \cos{\phi} \sin{\theta} & r \sin{\phi} \cos{\theta} \\
                              0 &           -r \sin{\theta} \\
      - \cos{\phi} \sin{\theta} &              - \sin{\phi} \\
      - \sin{\phi} \sin{\theta} &                \cos{\phi} \\
               1 - \cos{\theta} &                         0 \\
    \end{array}
    \right].
    \label{EQ::Jta_virtual}
\end{equation}

Then, the Jacobian matrix~${\bf J}_2(q_1,q_2) \in \mathbb{R}^{2 \times 2}$, which maps the motors velocities~$\left[\dot{q}_1,\dot{q}_2 \right]^\top$ to the angular velocities~$[\dot{\phi},\dot{\theta}]^\top$, is obtained upon time differentiation of Eq.~\eqref{EQ::a&t_1j}, and takes the form
\begin{equation}
	[\dot{\phi},\dot{\theta}]^\top =  {\bf J}_2(q_1, q_2)\left[\dot{q}_1,\dot{q}_2 \right]^\top\:\:\:\text{with}\:\:\:
    {\bf J}_2(q_1, q_2) = \frac{1}{2} \begin{bmatrix}
    1 & 1 \\
    -c & c \\
    \end{bmatrix}\:\:\:\text{where:}\:\:\: c = \frac{2 \tan{\alpha} \cos\left(\displaystyle\frac{q_1-q_2}{2}\right)}{1+\tan^2{\alpha} \sin^2\left(\displaystyle\frac{q_1-q_2}{2}\right)}.
    \label{EQ::taJq}
\end{equation}
The complete kinematic Jacobian matrix~${\bf J}$ of the NB-module is computed as
\begin{equation}
    {\bf J} = {\bf J}_1\; {\bf J}_2.
    \label{EQ::relation_TA}
\end{equation}
Then, the end-effector twist~$\mathbf{t}$ is expressed as a function of the motor velocities~$\left[\dot{q}_1,\dot{q}_2 \right]^\top$ as follows
\begin{equation}
    \mathbf{t} = {\bf J}\;\left[\dot{q}_1,\dot{q}_2 \right]^\top.
    \label{EQ::kinematic_Jacobian}
\end{equation}

\subsection{Module workspace}
\begin{figure}[!b]
\centering
\begin{minipage}{.49\textwidth}
   	\centering
    \includegraphics[scale=0.4]{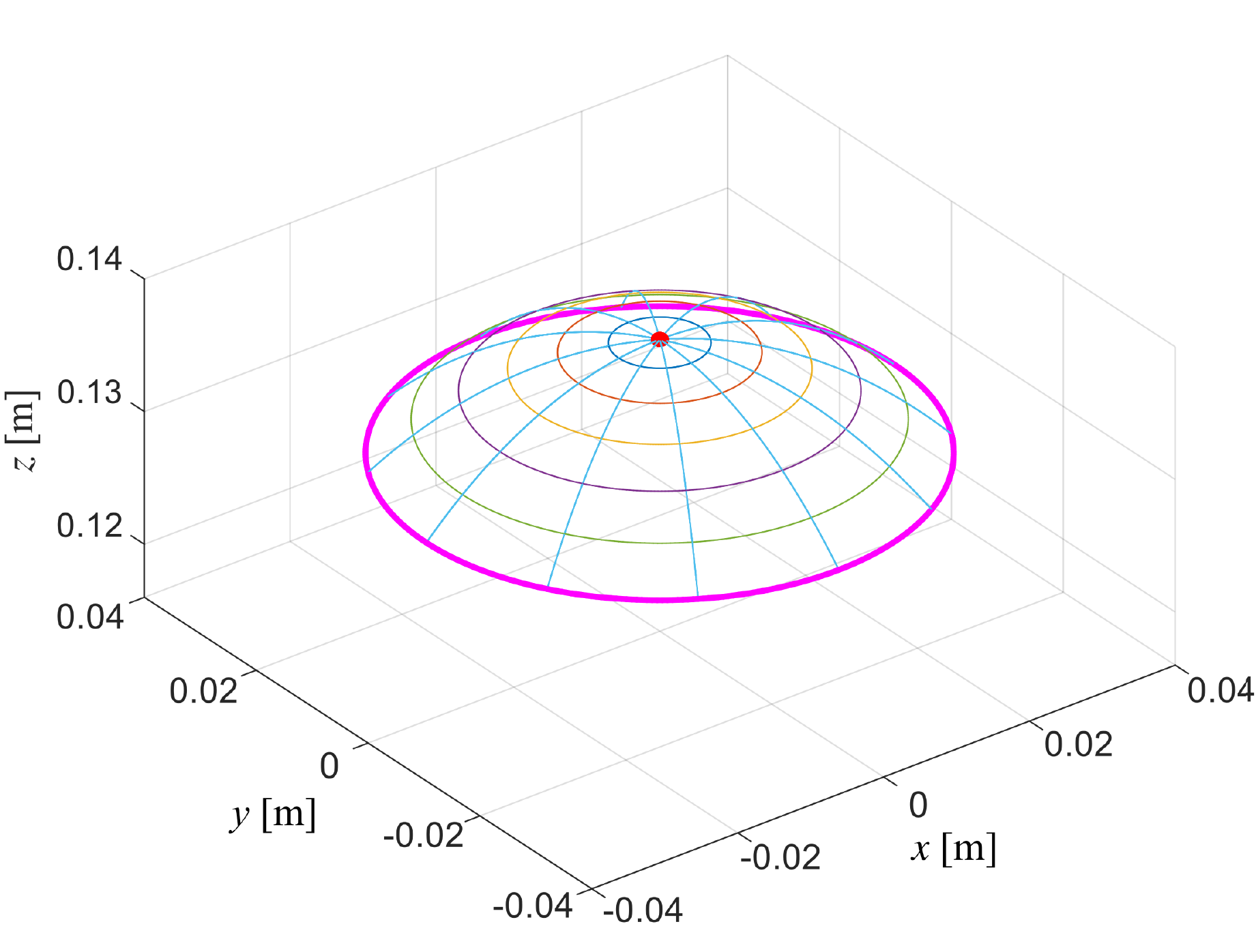}
    \caption{3D view of the workspace of one NB-module}
    \label{FIG::3d_ws}
\end{minipage}%
\hfill
\begin{minipage}{.49\textwidth}
 	\centering
    \includegraphics[scale=0.5]{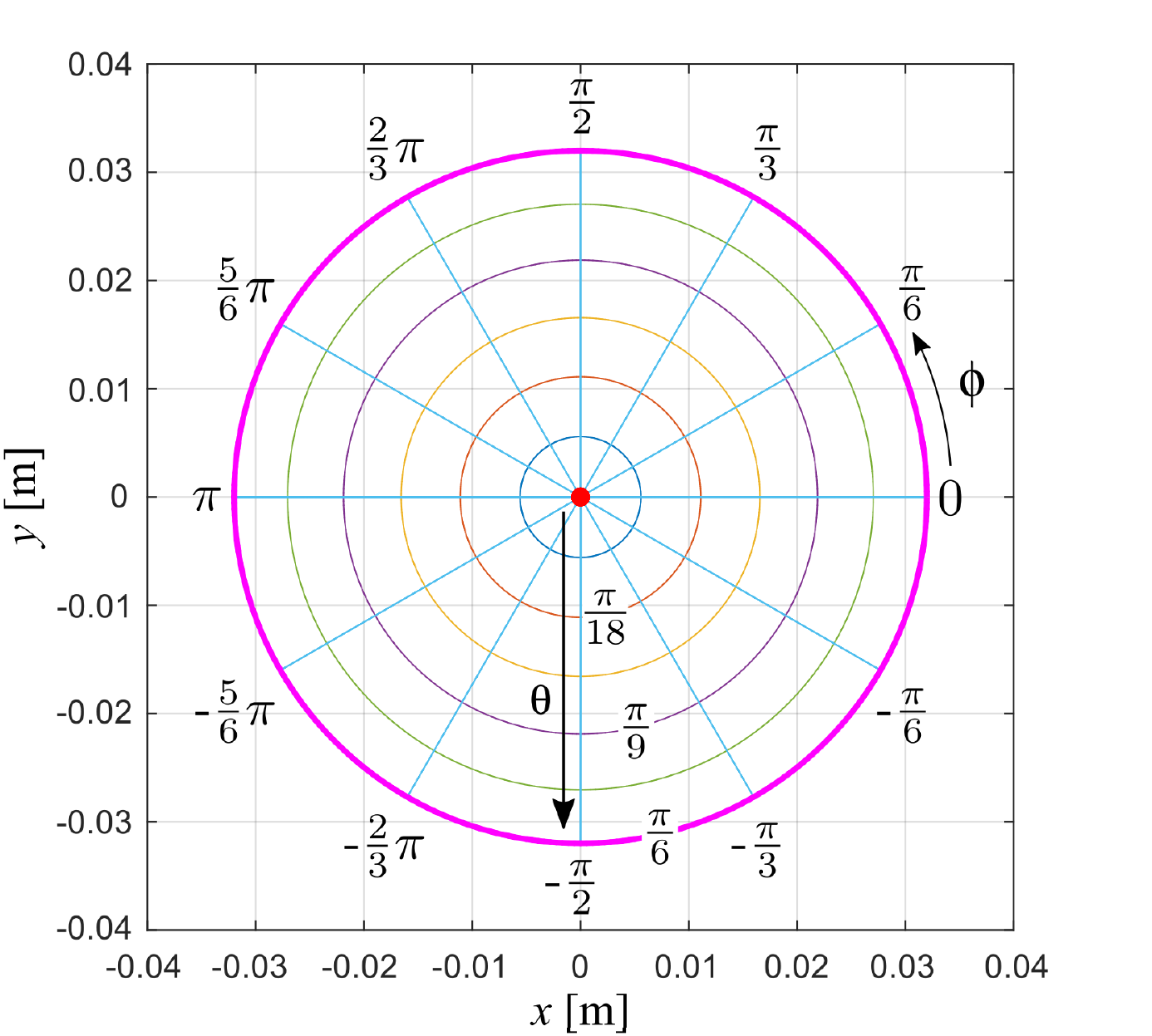}
    \caption{2D view of the workspace of one NB-module}
    \label{FIG::2d_ws}
\end{minipage}%
\end{figure}

The workspace of the NB-module is a portion of a sphere whose dimension depends on the length of~$r$ and the amplitude of~$2\alpha$.
Here the design parameters are set to~$r$~$=$~$0.07$~m and~$\alpha$~$=$~$\pi/12$~$=$~$15\degree$ to give an example of the NB-module workspace.
Figure~\ref{FIG::3d_ws} shows the~3D view of the workspace.
It corresponds to all the positions reached by frame~$\mathcal{F}_3$ for all the possible values of~$q_1$ and~$q_2$ in~$\left[ -\pi, \pi \right]$.
Each point on the sphere portion can be reached by two combinations of~$q_1$ and~$q_2$, both of them corresponding to the same orientation of the moving platform.

Figure~\ref{FIG::2d_ws} plots the tilt~$\theta$ and azimuth~$\phi$ angles on the~2D view of the workspace.
The value of~$\phi$ stays in the range~$[-\pi,\pi]$ and~$\theta$ in~$[-\pi / 6,\pi / 6]$.
In fact, the maximum absolute possible value of~$\theta$ is twice the slope of each tube, i.e.~$2\alpha$~$=$~$\pi / 6$.
The workspace of the NB-module is symmetric with respect to the \textit{z}-axis.

\section{Kinematic Control Algorithm}
\label{SEC::tpik}

This section describes the kinematic control algorithm used to control the RP-120.
Before introducing the kinematic control algorithm, some definitions are recalled from~\cite{Simetti2018priority}.
The vector~$\mathbf{q}\in\mathbb{R}^n$ is the joint variable vector, describing the arm configuration, where~$n$ is the number of joints.
The joint velocities are collected in the vector~$\dot{\mathbf{q}}\in \mathbb{R}^n$.

The notion of control objectives defines the goals of the robot.
A control objective is a scalar variable~$x(\mathbf{q})$ computed as a function of the robot configuration vector~$\mathbf{q}$ and represents the state of one task.
A control objective can be of two different types, equality and inequality.
Equality control objectives aim to satisfy the relationship~$x(\mathbf{q})$~$=$~$x_0$.
Inequality control objectives take the form~$x(\mathbf{q})\leq x_M$, or~$x(\mathbf{q}) \geq x_m$, or both simultaneously, where~$x_m$ and~$x_M$ are the lower and upper bounds of the variable~$x(\mathbf{q})$~\cite{Simetti2018priority}.
Control objectives can be divided into categories depending on their scope: system safety objectives, e.g. joint limits or obstacle avoidance, action oriented objectives, e.g. reaching a desired pose or following a desired trajectory, and optimization objectives, e.g. minimizing the joint velocities or optimizing the kinetostatic performance metrics.
This division is purely semantic and helps identify the correct priority level for each control objective.
Then, each scalar control objective is associated with a feedback reference rate~$\dot{\overline{x}}$.
The closed-loop rate control law drives the actual variable~$x(\mathbf{q})$ to the desired point~$x^*$ with the associated feed-forward changing rate~$\dot{x}^*$ and is defined as
\begin{equation}
  \dot{\overline{x}}= \lambda(x^*-x(\mathbf{q}))+\dot{x}^*,
\end{equation}
where~$\lambda$ is a positive gain related to the target convergence rate.
The actual derivative of~$x$ is defined as a function of the joint velocity vector~$\dot{\mathbf{q}}$ as follows:
\begin{equation}
  \dot{x}(\mathbf{q},\dot{\mathbf{q}})=\mathbf{J}_{\text{task}}(\mathbf{q})\dot{\mathbf{q}}=\begin{bmatrix}\frac{\partial x}{\partial q_1}&\dots&\frac{\partial x}{\partial q_n}\end{bmatrix}\dot{\mathbf{q}},
\end{equation}
where~$\mathbf{q}$~$=$~$[q_1\;\dots\;q_n]$.

An activation function~$a^i(x) \in [0,1]$ is associated to each control objective~$x(\mathbf{q})$, and it represents whether the objective is relevant or not in a given time instant.
The tasks associated with inequality control objectives are relevant only when the scalar variable~$x(\mathbf{q})$ is near or out of the validity region.
So, the activation function assumes zero values within the validity region of the associated inequality objective and one when it is not, with a smooth transition between the two states.
For tasks associated with equality control objectives, the activation function is set to~$a^i(x) = 1$ because they always need to be active.

A specific priority is assigned to each task based on the relative importance of each objective.
The meaning of the priority is that the highest priority tasks are solved first using the available robot degrees-of-freedom and are not affected by the lower priority ones.
Hence, lower priority tasks are solved if enough robot degrees-of-freedom remains.
When two or more tasks have the same priority, they are grouped in a multidimensional control task.
A specific list of prioritized tasks is called control action~$\mathscr{A}$.

With the previous definitions, the following quantities associated with each priority level in a control action~$\mathscr{A}$ can be defined~\cite{Simetti2019priority}:
\begin{itemize}
  \item $\dot{\overline{\mathbf{x}}}_k$~$=$~$[\dot{\overline{x}}_{1,k},\;\dot{\overline{x}}_{2,k},\;\dots\;,\;\dot{\overline{x}}_{m_k,k}]^\top$ is the vector collecting all the reference rates of the scalar control tasks, where~$m_k$ is the number of scalar tasks for the priority level~$k$.
  \item $\mathbf{J}_k$ is the Jacobian matrix associated with the~$k^{\text{th}}$ task vector~$[\dot{x}_{1,k},\;\dots\;,\;\dot{x}_{m_k,k}]^\top$ with respect to the joint velocity vector~$\dot{\mathbf{q}}$.
  \item $\mathbf{A}_k$~$=$~$\text{diag}(a_{1,k},\;\dots\;,\;a_{m_k,k})$ is a diagonal matrix of the activation functions.
\end{itemize}

To find the system velocity reference vector~$\dot{\overline{\mathbf{q}}}$ that meets the priority requirements of a given action, the TPIK algorithm solves a sequence of nested minimization problems
\begin{equation}
  S_k=\text{arg}\;\underset{\dot{\overline{\mathbf{q}}}\in S_{k-1}}{\text{R}-\text{min}} ||\mathbf{A}_k(\dot{\overline{\mathbf{x}}}_k - \mathbf{J}_k\dot{\overline{\mathbf{q}}})||^2,
  \label{EQ::tpik_solution}
\end{equation}
where~$S_{k}$ is the solution at the~$k^\text{th}$ task and~$S_{k-1}$ is the manifold of solutions of the previous priority level.
The TPIK algorithm uses regularized space projection to implement priorities among the tasks.
These computations are highlighted by the notation~$\text{R}-\text{min}$.
In addition to the minimization performed in Eq.~\eqref{EQ::tpik_solution}, other regularization costs are included.
These regularization costs are necessary to avoid discontinuities in the system velocity vector due to kinematic and algorithmic singularities.
These regularization costs will not be analyzed in this work, since they are fully explained in~\cite{Simetti2016inequality}.

A significant advantage of the TPIK algorithm is the use of the activation functions to handle inequality control objectives without over-constraining the system.
Both equality and inequality control require a certain amount of robot degrees-of-freedom specified by the associated task.
When an inequality task is inside its validity region, the activation function goes to zero, therefore not consuming any degrees-of-freedom.
So, safety tasks, like joint limits, can be placed at the top of the hierarchy without over-constraining the system.

Finally, the TPIK algorithm adopts another continuous sigmoidal function~$a^p(\mathbf{p})$ which includes the previous and current executed actions and the time elapsed in the current step to perform a smooth activation/deactivation transition between two actions.
This function~$a^p(\mathbf{p})$ is used together with ~$a^i(x)$.
More details are presented in~\cite{Simetti2018priority}.

\section{Kinetostatic Tasks for Robot Performance Optimization}
\label{SEC::kinetostatic_metrics}

The TPIK algorithm considers two tasks related to the end-effector pose and velocity to control the RP-120 and performs a series of machining tasks.
The kinematic Jacobian matrix~$\mathbf{J}_{e}$ of these tasks is the one that relates the end-effector twist~$\mathbf{t}\in\mathbb{R}^6$ to the joint velocity vector~$\dot{\mathbf{q}}\in\mathbb{R}^n$,
\begin{equation}
  \mathbf{t}=
  \begin{bmatrix}\dot{\mathbf{p}}\\ \bm{\omega}\end{bmatrix}=
  \mathbf{J}_{e}(\mathbf{q})\dot{\mathbf{q}}=
  \begin{bmatrix}\mathbf{J}_l(\mathbf{q})\\ \mathbf{J}_a(\mathbf{q})\end{bmatrix}\dot{\mathbf{q}}.
  \label{EQ::twist}
\end{equation}

Other two tasks, based on the robot kinetostatic performance metrics, are used to optimize the robot performance.
This section describes the two kinetostatic performance metrics that rate the kinetostatic robot abilities.
Together with the definition of these metrices, their Jacobian matrices are computed as a function of the robot configuration vector~$\mathbf{q}$.
In this way, the TPIK algorithm uses these kinetostatic tasks to optimize the robot performance while performing the machining operations.
The kinetostatic performance metrics proposed in this paper are the dexterity~\cite{Angeles1992kinematic} and the robot transmission ratio~(RTR)~\cite{zargarbashi2012posture}.

Before introducing the definitions of dexterity and RTR, it is essential to recall that the kinematic Jacobian matrix needs to be weighted to compute these metrics since it contains non-homogeneous terms, i.e. linear and angular ones.
The weighting of~$\mathbf{J}_{e}$ employs the characteristic length~$L$.
It was introduced in \cite{angeles1992design} to solve the absence of dimensional homogeneity in the kinematic Jacobian matrix entries and its determination is described in~\cite{khan2006kinetostatic}.
To weight~$\mathbf{J}_{e}$, the revolute joint columns of the linear kinematic Jacobian matrix part are divided by~$L$.
The weighted kinematic Jacobian matrix of the RP-120, which accounts only revolute joints, is written as
\begin{equation}
\mathbf{J}_{w}(\mathbf{q})=\begin{bmatrix}\mathbf{J}_l(\mathbf{q})/L\\ \mathbf{J}_a(\mathbf{q})\end{bmatrix}.
\end{equation}
\subsection{Dexterity}

The dexterity~$\eta_1(\mathbf{J}_{w})$ characterizes the kinematic performance of a manipulator in a given configuration and is defined as the inverse of the conditioning number~$\kappa(\mathbf{J}_{w})$ of its Jacobian matrix~\cite{Angeles1992kinematic}:
\begin{equation}
    \kappa(\mathbf{J}_{w}) = ||\mathbf{J}_{w}||\:||\mathbf{J}_{w}^{-1}||\:\:\:\text{and}\:\:\:\eta_1(\mathbf{J}_{w}) = 1 / \kappa(\mathbf{J}_{w}).
    \label{EQ::cond_num_dxt}
\end{equation}
The index~$\eta_1$ is bounded by~0 and~1.
The higher~$\eta_1$, the better the manipulator dexterity.
If~$\eta_1$~$=$~$1$, the robot is in an isotropic configuration.
The smaller~$\eta_1$, the worse the manipulator dexterity and the closer to a singularity.
Moreover,~$\eta_1$ corresponds to the ratio between the smallest and highest singular values of~$\mathbf{J}_{w}$ denoting how close the manipulability hyper-ellipsoid is to being a hyper-sphere~\cite{pond2006formulating}.
So, in case of~$\eta_1$~$=$~$1$, the robot can move with the same velocity amplification factor in all the task space directions.
Contrary, in case of~$\eta_1$~$=$~$0$, the robot cannot move in one or more task space directions.

In this case, the Frobenius norm of~$\mathbf{J}_{w}$~\cite{rakotomanga2008kinetostatic} is employed to obtain an analytical expression of~$\eta_1$:
\begin{equation}
    \eta_1(\mathbf{J}_w) = \frac{m}{\sqrt{\text{trace}(\mathbf{J}_w \mathbf{J}_w^\top )\:\text{trace}[(\mathbf{J}_w\mathbf{J}_w^\top )^{-1}]}},
    \label{EQ::dxt_frobenius}
\end{equation}
where~$m$ is the number of rows of~$\mathbf{J}_{w}$ and represents the dimension of task space.
The following definitions are introduced to increase the readability of the equations:
\begin{equation}
	\gamma_1(\mathbf{J}_w)\triangleq\sqrt{\text{trace}(\mathbf{J}_w \mathbf{J}_w^\top )}\:\:\:\text{and}\:\:\:\gamma_2(\mathbf{J}_w)\triangleq\sqrt{\text{trace}[(\mathbf{J}_w\mathbf{J}_w^\top )^{-1}]}.
\end{equation}
where~$\triangleq$ represents the definition symbol.
So,~$\eta_1$ can be rewritten as
\begin{equation}
	\eta_1(\mathbf{J}_w) = \frac{m}{\gamma_1(\mathbf{J}_w)\:\gamma_2(\mathbf{J}_w)}.
\end{equation}

The dexterity Jacobian matrix is determined to relate the velocity rate of~$\eta_1$ with respect to the joint velocity vector~$\dot{\mathbf{q}}$.
Since the Frobenius formula used in Eq.~\eqref{EQ::dxt_frobenius} expresses~$\eta_1$ as a function of joint position vector~$\mathbf{q}$ in an analytical way, it allows its derivation.
So, the derivative of Eq.~\eqref{EQ::dxt_frobenius} with respect to each joint position~$q_i\in\mathbf{q}$ is
\begin{equation}
\begin{split}
\frac{\partial \eta_1}{\partial q_i}=-\eta_1
\Bigg(\frac{\partial \gamma_1}{\partial q_i}\frac{1}{\gamma_1}
+\frac{1}{\gamma_2}\frac{\partial \gamma_2}{\partial q_i}\Bigg),
\end{split}
\label{EQ::dxt_deriv}
\end{equation}
where
\begin{equation}
\frac{\partial \gamma_1}{\partial q_i}=\frac{1}{\gamma_1}\text{trace}\Bigg\{\mathbf{J}_{w}\frac{\partial\mathbf{J}_{w}^\top}{\partial q_i}\Bigg\}\:\:\:\text{and}\:\:\:
\frac{\partial \gamma_2}{\partial q_i}
=\frac{1}{\gamma_2}
\text{trace}
\Bigg\{-\mathbf{J}_{w}
\frac{\partial\mathbf{J}_w^\top}{\partial q_i}
(\mathbf{J}_w\mathbf{J}_w^\top )^2
\Bigg\}.
\end{equation}

In conclusion, the dexterity Jacobian matrix~$\mathbf{J}_{\eta_1}$ as a function of the joint variables is
\begin{equation}
\mathbf{J}_{\eta_1}=\begin{bmatrix}
\frac{\partial \eta_1}{\partial q_1}&\dots&\frac{\partial \eta_1}{\partial q_n}
\end{bmatrix},
\end{equation}
where~$n$ is the number of columns of~$\mathbf{J}_{w}$ and represents the dimension of joint space.


\subsection{Robot transmission ratio}

The RTR~$\eta_2(\mathbf{J}_{w})$ measures how effectively the actuator forces produce the desired robot motion~\cite{zargarbashi2012posture}.
It corresponds to the angle between the joint velocity~$\dot{\mathbf{q}}$ and torque~$\bm{\tau}$ vectors
\begin{equation}
	\eta_2 =\frac{|\bm{\tau}^\top\dot{\mathbf{q}}|}{||\bm{\tau}||\:||\dot{\mathbf{q}}||}=|\cos\angle(\bm{\tau},\dot{\mathbf{q}})|.
\end{equation}
This index is bounded between~0 and~1.
Maximizing~$\eta_2$ protects the joint motors not only from velocity saturation but also from torque saturation~\cite{zargarbashi2012posture}.
In case of kinetostatic redundancy,~$\eta_2$ can also be written as a function of the end-effector twist~$\mathbf{t}$ and the wrench~$\mathbf{w}$ applied to it:
\begin{equation}
   \eta_2 =\frac{|\mathbf{w}^\top\mathbf{t}|}{||\mathbf{J}_{w}^\top \mathbf{w}||\:||\mathbf{J}_{w}^+\mathbf{t}||},
   \label{EQ::RTR}
\end{equation}
where~$\mathbf{t}$ is the robot end-effector twist defined in Eq.~\eqref{EQ::twist} and~$\mathbf{w}$~$=$~$[\mathbf{f}^\top,\mathbf{m}^\top]^\top$ is the wrench that collects the forces~$\mathbf{f}$ and moments~$\mathbf{m}$ exerted by the environment on the end-effector.
The matrix~$\mathbf{J}_{w}^+$ is the pseudo-inverse of the weighted kinematic Jacobian matrix.
To ensure that~$\eta_2$ is dimensionless, the linear part~$\dot{\mathbf{p}}$ in~$\mathbf{t}$ and the moment~$\mathbf{m}$ in~$\mathbf{w}$ are divided by the characteristic length~$L$.

The~RTR Jacobian matrix is obtained by doing the derivative of Eq.~\eqref{EQ::RTR} with respect to each joint position~$q_i\in\mathbf{q}$:
\begin{equation}
\frac{\partial \eta_2}{\partial q_i} = \eta_2\;\frac{\mathbf{w}^\top\mathbf{J}_{w}\frac{\partial \mathbf{J}_{w}^\top}{\partial q_i}\mathbf{w}||\mathbf{J}_{w}^+\mathbf{t}||^2-||\mathbf{J}_{w}^\top\mathbf{w}||^2\mathbf{t}^\top\mathbf{J}_{w}^{+^\top}\frac{\partial \mathbf{J}_{w}^+}{\partial q_i}\mathbf{t}}{(||\mathbf{J}_{w}^\top\mathbf{w}||\; ||\mathbf{J}_{w}^+\mathbf{t}||)^2},
\end{equation}
where the values of the end-effector twist~$\mathbf{t}$ and the wrench~$\mathbf{w}$ applied to it are defined in the trajectory planning and therefore known.
The derivative of the pseudo-inverse weighted kinematic Jacobian matrix~$\partial \mathbf{J}_{w}^+/\partial q_i$ is defined in~\cite{golub1973differentiation}.

The~RTR Jacobian matrix~$\mathbf{J}_{\eta_2}$ as a function of the joint variables is
\begin{equation}
\mathbf{J}_{\eta_2}=\begin{bmatrix}
\frac{\partial\eta_2}{\partial q_1}&\dots&\frac{\partial\eta_2}{\partial q_n}
\end{bmatrix},
\end{equation}
where~$n$ is the number of columns of~$\mathbf{J}_{w}$ and represents the dimension of joint space.


\section{Trajectory Tracking with the Nimbl'Bot Robot}
\label{SEC::test}

\begin{figure}[!t]
    \centering
    \begin{minipage}{.49\textwidth}
	   	\centering
    	 \includegraphics[scale=0.59]{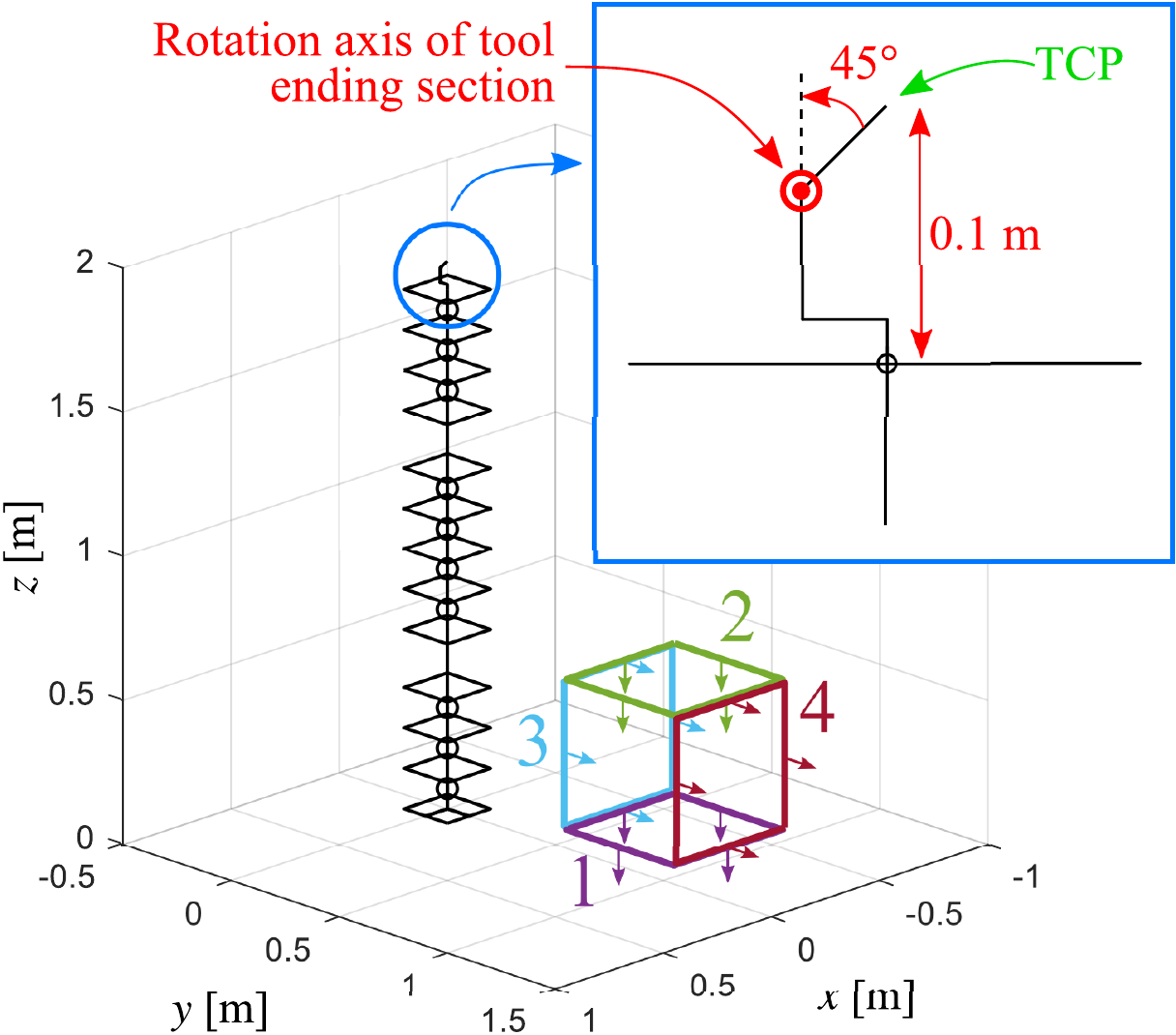}
	    \caption{Schematics of the RP-120 with the four trajectories to track in 3D space. In the top right corner the machining tool attached to the RP-120 end-effector used in the cutting phase. Tool center point (TCP) highlighted in green. Tool ending section rotated around red point of~$45^{\circ}$.}
    	\label{FIG::robot_and_trajectories}
    \end{minipage}%
	\hfill
	\begin{minipage}{.49\textwidth}
 		\centering
   	    \includegraphics[scale=0.45]{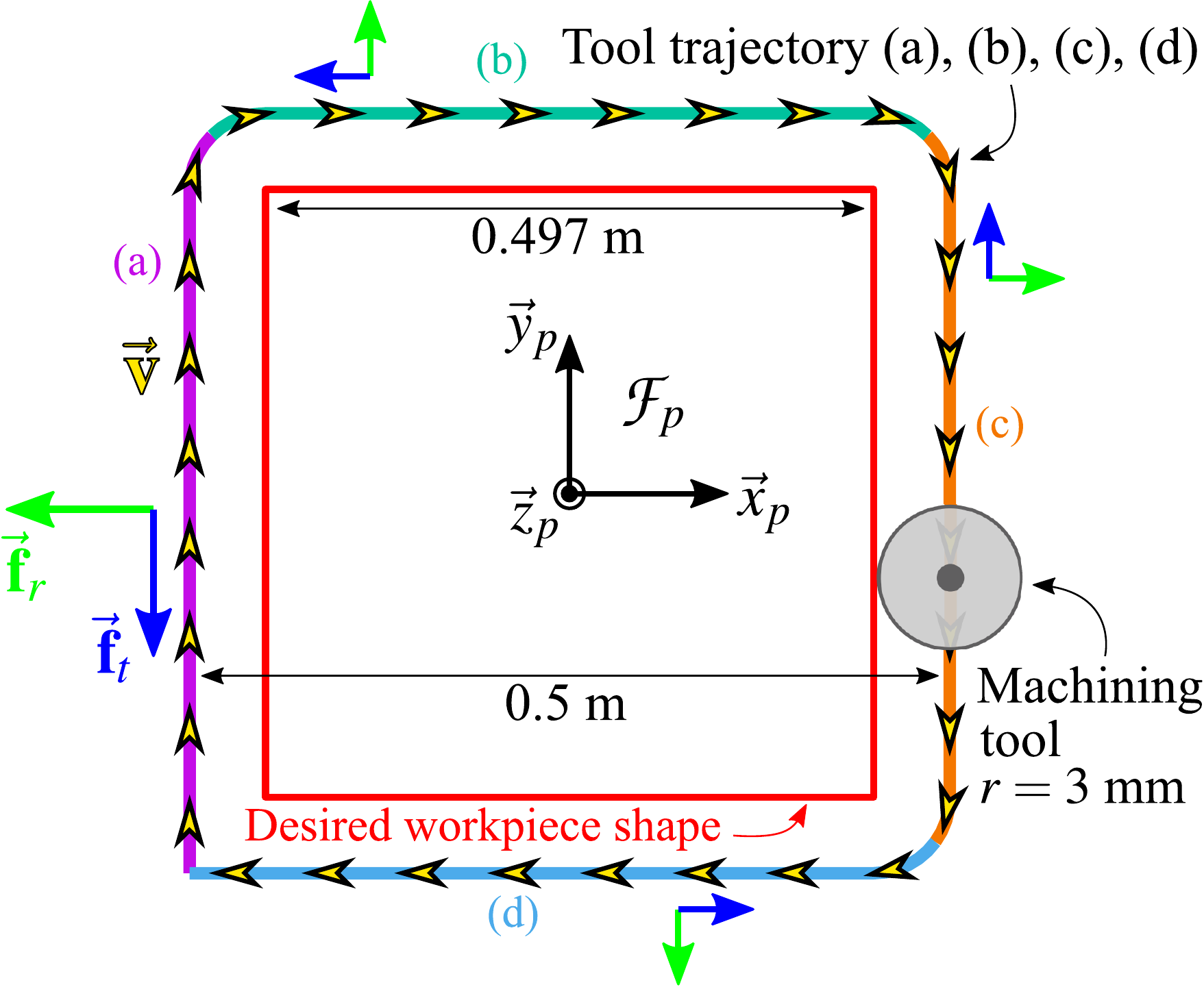}
    	\caption{Measures of the desired workpiece, machining tool and trajectory tool with workpiece frame~$\mathcal{F}_p$. Orientation of velocity vector~$\vec{\mathbf{v}}$ (yellow) plus tangential and radial force vectors~$\vec{\mathbf{f}}_t$ and~$\vec{\mathbf{f}}_r$ (blue and green). The tool trajectory is divided into four parts~(a),~(b),~(c) and~(d).}
    	\label{FIG::trajectory_schematics}
	\end{minipage}%
	\hfill
    \begin{minipage}{.99\textwidth}
 		\centering
    	 \includegraphics[scale=0.5]{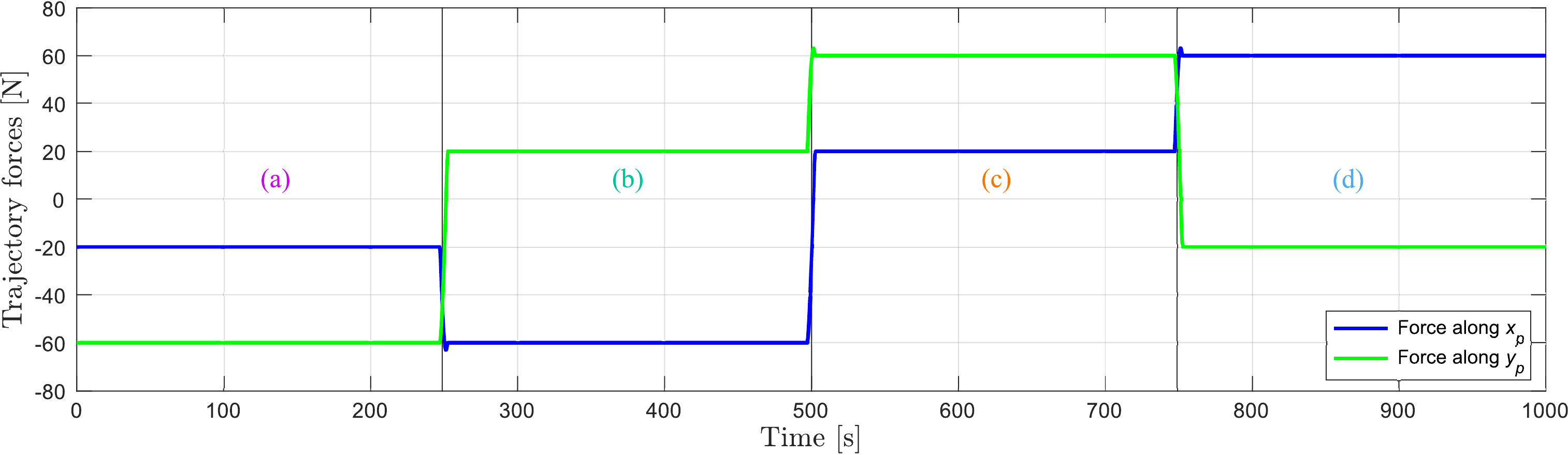}
    	\caption{Velocity profiles of the machining tool in frame~$\mathcal{F}_{p}$. Each sector is labeled (a), (b), (c) and (d) to match the corresponding part of the trajectory.}
    	\label{FIG::trajectory_velocities}
	\end{minipage}%
	\hfill
	\begin{minipage}{.99\textwidth}
	   	\centering
    	 \includegraphics[scale=0.5]{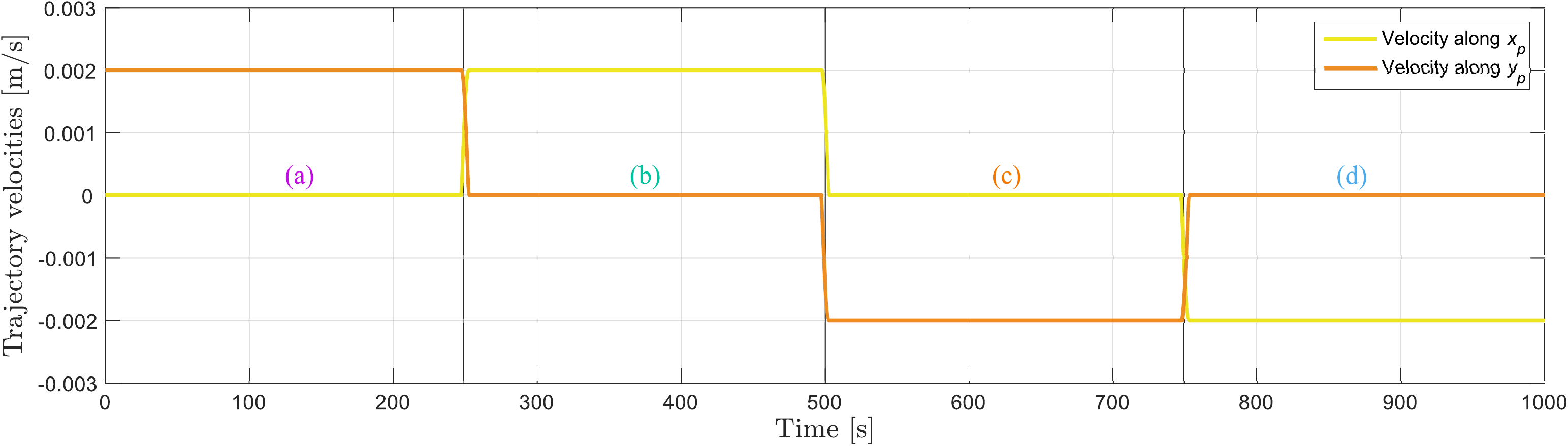}
	    \caption{Cutting force profiles in frame~$\mathcal{F}_{p}$. Each sector is labeled (a), (b), (c) and (d) to match the corresponding part of the trajectory.}
    	\label{FIG::trajectory_forces}
    \end{minipage}%
\end{figure}

This section describes the tests performed in a computer simulation on the RP-120 design and discusses the obtained results.
The test consists in a joint trajectory planning and makes the RP-120 track different trajectories with and without activating the tasks related to dexterity and RTR.
After collecting the metric values on each trajectory, these are compared to demonstrate the benefit of using the optimization tasks and identify the best joint trajectory planning for this robot on each trajectory.
The RP-120 has to track four trajectories of the same shape and size.
Figure~\ref{FIG::robot_and_trajectories} shows the RP-120 next to the four trajectories.
Two of them are oriented horizontally and the others are vertical.
These trajectories describe a cubic area whose side are 0.5~m~$\times$~0.5~m centered in~$(x,y,z)$~$=$~$(0.0,1.05,0.45)$.
The machining tool is shown in the top right corner of Fig.~\ref{FIG::robot_and_trajectories}.
The tool ending part is rotated of~$45^{\circ}$ around the red point.
This allows the robot to reach all the points on each trajectory.
The trajectories are planned to cut a squared shape using a machining tool and the measures are shown in Fig.~\ref{FIG::trajectory_schematics}.
The tool trajectory is divided in four parts~(a),~(b),~(c) and~(d).
Figure~\ref{FIG::trajectory_schematics} also shows the orientation of the velocity vector~$\vec{\mathbf{v}}$ and the tangential and radial force vectors~$\vec{\mathbf{f}}_t$ and~$\vec{\mathbf{f}}_r$ applied on the machining tool.
The magnitudes of~$\vec{\mathbf{v}}$,~$\vec{\mathbf{f}}_t$ and~$\vec{\mathbf{f}}_r$ are constant along the entire trajectory.
The profiles of~$\vec{\mathbf{v}}$,~$\vec{\mathbf{f}}_t$ and~$\vec{\mathbf{f}}_r$ are depicted in Figs.~\ref{FIG::trajectory_velocities} and~\ref{FIG::trajectory_forces}.
The gravity force and the cutting one along~$\vec{z}_p$ are neglected in this work.
The details about the RP-120 and trajectory features are in Tables~\ref{TAB::robot_details} and~\ref{TAB::trajectory_details}, respectively.
The details of the machine and the implementation are given in Table~\ref{TAB::simulation_details}.
The RP-120 features, trajectory size and velocity/force magnitudes were provided by Nimbl'Bot.
\begin{table}[!b]
\centering
\begin{minipage}[t]{.49\textwidth}
\centering
\caption{Main robot dimensions plus joint velocity and acceleration limits}
\label{TAB::robot_details}
\begin{tabular}{ c c }
\hline\hline
Module half height $r$ & $0.07$ m \\
Module tube slope $\alpha$ & $15^{\circ}$ \\
Link length & $0.2$ m \\
Tool height & $0.1$ m \\
Tool offset & $45^{\circ}$ \\
Robot + tool total height & $1.9$ m \\
Max/min joint velocity & $\pm 1.0$ rad/s \\
Max joint acceleration/deceleration & $2.0$ rad/s$^2$ \\\hline\hline
\end{tabular}
\end{minipage}%
\hfill
\begin{minipage}[t]{.49\textwidth}
\centering
\caption{Test trajectory details, velocities and forces exerted on end-effector and time for tracking entire trajectory}
\label{TAB::trajectory_details}
\begin{tabular}{ c c }
\hline\hline
Square side & $0.5$ m \\
Steps & 2001 \\
Magnitude velocity vector $||\vec{\mathbf{v}}||$ & $0.002$ m/s \\
Magnitude tangential force vector $||\vec{\mathbf{f}}_t||$ & $60$ N \\
Magnitude radial force vector $||\vec{\mathbf{f}}_r||$ & $20$ N \\
Time & $1000$ s \\\hline\hline
\end{tabular}
\end{minipage}%
\end{table}
\begin{table}[!b]
\centering
\caption{Machine and implementation details}
\label{TAB::simulation_details}
\begin{tabular}{ c c }
\hline\hline
Operating System & Linux \\
Distribution & Ubuntu 20.04 \\
CPUs number & 4 \\
CPU model & Intel Core i7 10th Gen, 1.30GHz \\
Language & C++ \\
Control frequency & 10Hz\\
Total simulation time & 2.5 hours \\\hline\hline
\end{tabular}
\end{table}

Since both dexterity~$\eta_1$ and RTR~$\eta_2$ are employed, they are combined in a linear function used to rate the performance of the RP-120 on each trajectory,
\begin{equation}
\eta = \lambda_1\eta_1+\lambda_2\eta_2,
\label{EQ::linear_relationship}
\end{equation}
where~$\lambda_1$ and~$\lambda_2$ are scaling factors.
Since~$\eta_1$ and~$\eta_2$ are valid in the same range, the weighting factors are selected as~$\lambda_1$~$=$~$\lambda_2$~$=$~$0.5$ and~$\eta$ becomes valid in the same range~$[0,1]$
So,~$\eta=1$ means that the robot is in an isotropic configuration and the angle between the joint velocity and torque vectors tends to~$0^{\circ}$.

Four different actions are used in the simulation.
When the dexterity and RTR tasks are deactivated,~$\mathscr{A}_1$ is the action used to reach the starting pose and~$\mathscr{A}_2$ to follow the trajectory.
When the optimization tasks are activated,~$\mathscr{A}_3$ brings the robot to the starting pose and~$\mathscr{A}_4$ follows the trajectory.
Table~\ref{TAB::tasks_hierarchy} shows the task details and their hierarchy inside each action.
\begin{table}[!t]
\centering
\caption{Details about the task names, control objective types, and hierarchy levels.
Symbol~(E) represents the equality control objective tasks and~(I) the inequality control objective tasks.
The last four columns list the hierarchy level for each task in actions~$\mathscr{A}_1$~(Reach Pose),~$\mathscr{A}_2$~(Follow Trajectory),~$\mathscr{A}_3$~(Reach Pose Optimized) and~$\mathscr{A}_4$~(Follow Trajectory Optimized). When symbol~(/) is used, it means that a task is not present in the action and has no hierarchy level.}
\begin{tabular}{ |c|c|c|c|c|c|c| }
 \cline{4-7}
 \multicolumn{2}{c}{} & & \multicolumn{4}{c|}{Hierarchy levels} \\\hline
 Task & Category & Type &~$\mathscr{A}_1$ &~$\mathscr{A}_2$ &~$\mathscr{A}_3$ &~$\mathscr{A}_4$ \\\hline
 End-Effector Pose & action oriented & E & $1^{\text{st}}$ & / & $1^{\text{st}}$ & / \\
 End-Effector Velocity & action oriented & E & / & $1^{\text{st}}$ & / & $1^{\text{st}}$ \\
 Dexterity & optimization & I & / & / & $2^{\text{nd}}$ & $2^{\text{nd}}$ \\
 RTR & optimization & I & / & / & $2^{\text{nd}}$ & $2^{\text{nd}}$ \\\hline
\end{tabular}
\label{TAB::tasks_hierarchy}
\end{table}
\begin{algorithm}[!t]
\caption{Procedure to compare kinetostatic optimization in machining operations}
\label{ALG::method}
\begin{algorithmic}
\Require Actions~$\mathscr{A}_1~=$~Reach Pose,~$\mathscr{A}_2~=$~Follow Trajectory,~$\mathscr{A}_3~=$~Reach Pose Optimized and~$\mathscr{A}_4~=$~Follow Trajectory Optimized.
\Variables Number of trajectories~($n_t$) is~$4$ and number of repetitions~($n_r$) is~$100$. Variable~$t$ is time step. Object~$robot$ represents the simulated design, contains joint position vector~$\mathbf{q}$ and values of~$(\eta,\eta_1,\eta_2)$, can use the TPIK function to perform the desired action~$\mathscr{A}$. Object~$trajectory$ contains all the poses, velocities and forces at each~$\mathbf{Step}(0:End)$.
\end{algorithmic}
\begin{algorithmic}[1]
\For{$i := 1\rightarrow$ $n_t$}
	\For{$j := 1\rightarrow$ $n_r$}
		\State $robot.\mathbf{q} = \text{RandomInit}()$ \Comment{Random initialization of the robot configuration} \label{ALG::random}
		\State $\text{Save}(robot.\mathbf{q})$ \Comment{Same random starting configuration reused with optimization tasks, step~\ref{ALG::reload}}
		\State Load $trajectory(i)$
		\While{$!trajectory.\mathbf{Step}(0)$} \Comment{Start machining task not optimized}
			\State $robot.\text{TPIK}(\mathscr{A}_1,trajectory.\mathbf{Step}(0))$ \Comment{Run~$\mathscr{A}_1$ with TPIK algorithm to reach starting pose}
		\EndWhile
		\While{$!trajectory.\mathbf{Step}(End)$}
			\State $robot.\text{TPIK}(\mathscr{A}_2,trajectory.\mathbf{Step}(t))$ \Comment{Run~$\mathscr{A}_2$ with TPIK algorithm to follow trajectory}
			\State $robot.\eta(t) = 0.5(robot.\eta_1(t)+robot.\eta_2(t))$
			\State $\text{Save}(robot.[\eta(t),\eta_1(t),\eta_2(t)])$ \Comment{Save performance metrics for result analysis}
		\EndWhile
		\State Reload saved~$robot.\mathbf{q}$ \Comment{Use same random starting configuration generated at step~\ref{ALG::random}} \label{ALG::reload}
		\While{$!trajectory.\mathbf{Step}(0)$ \& $robot.[\eta_1(t),\eta_2(t)] \neq$ local max} \Comment{Start machining task optimized}
			\State $robot.\text{TPIK}(\mathscr{A}_3,trajectory.\mathbf{Step}(0))$ \Comment{Run~$\mathscr{A}_3$ with TPIK algorithm to reach starting pose}
		\EndWhile \Comment{Stop if end-effector in desired pose and~$[\eta_1(t),\eta_2(t)]$ in local max}
		\While{$!trajectory.\mathbf{Step}(End)$}
			\State $robot.\text{TPIK}(\mathscr{A}_4,trajectory.\mathbf{Step}(t))$ \Comment{Run~$\mathscr{A}_4$ with TPIK algorithm to follow trajectory}
			\State $robot.\eta(t) = 0.5(robot.\eta_1(t)+robot.\eta_2(t))$
			\State $\text{Save}(robot.[\eta(t),\eta_1(t),\eta_2(t)])$ \Comment{Save performance metrics for result analysis}
		\EndWhile
	\EndFor
\EndFor
\end{algorithmic}
\end{algorithm}
\begin{figure}[!t]
\centering
\begin{minipage}[t]{.49\textwidth}
   	\centering
    \includegraphics[scale=0.54]{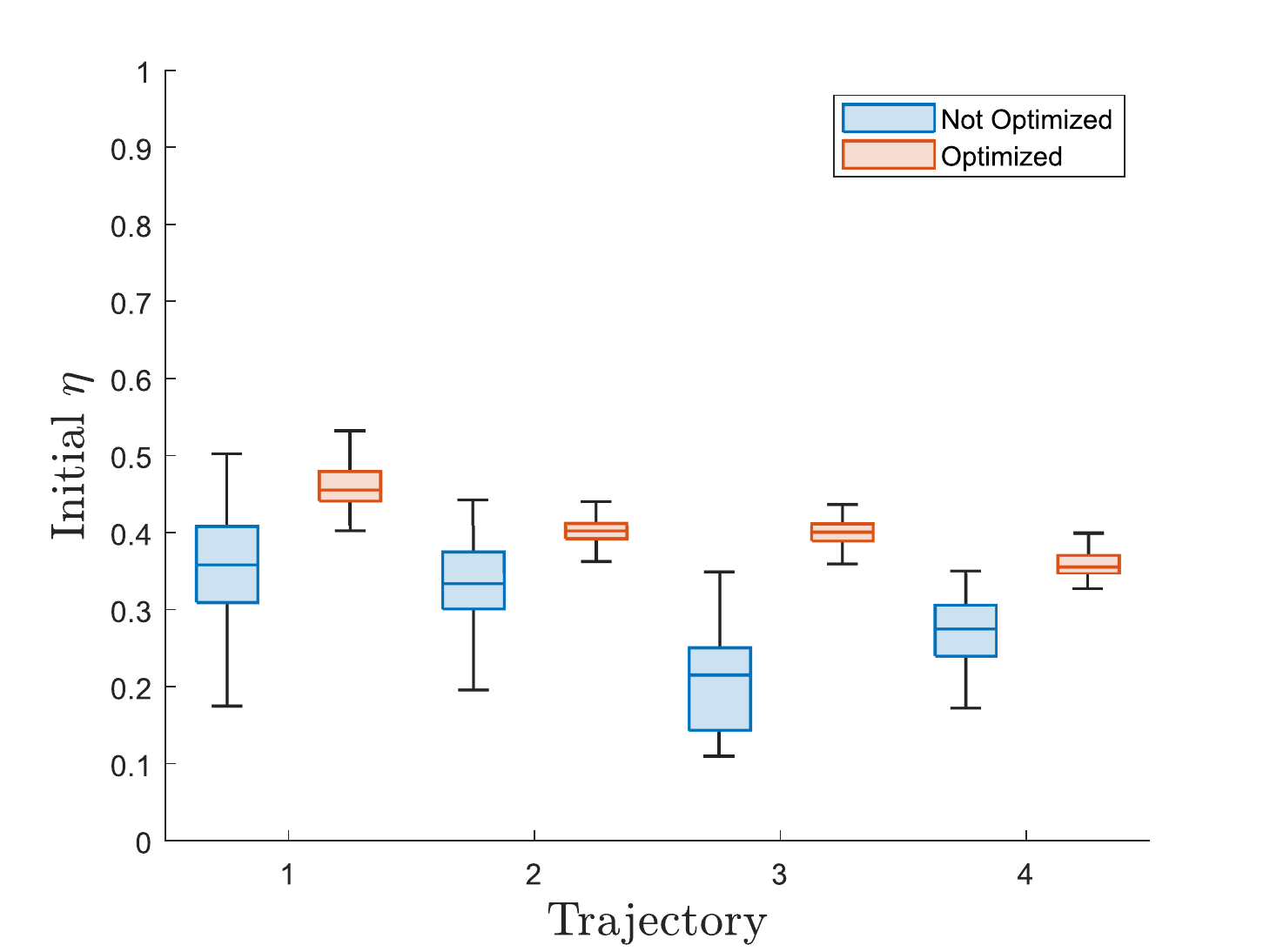}
    \caption{Values taken by~$\eta$ at the starting pose of each trajectory (1~to~4, Fig.~\ref{FIG::robot_and_trajectories}) for each one of the hundred repetitions, without (blue) and with (red) optimization of dexterity and RTR}
    \label{FIG::init_sigma}
\end{minipage}%
\hfill
\begin{minipage}[t]{.49\textwidth}
	\centering
    \includegraphics[scale=0.525]{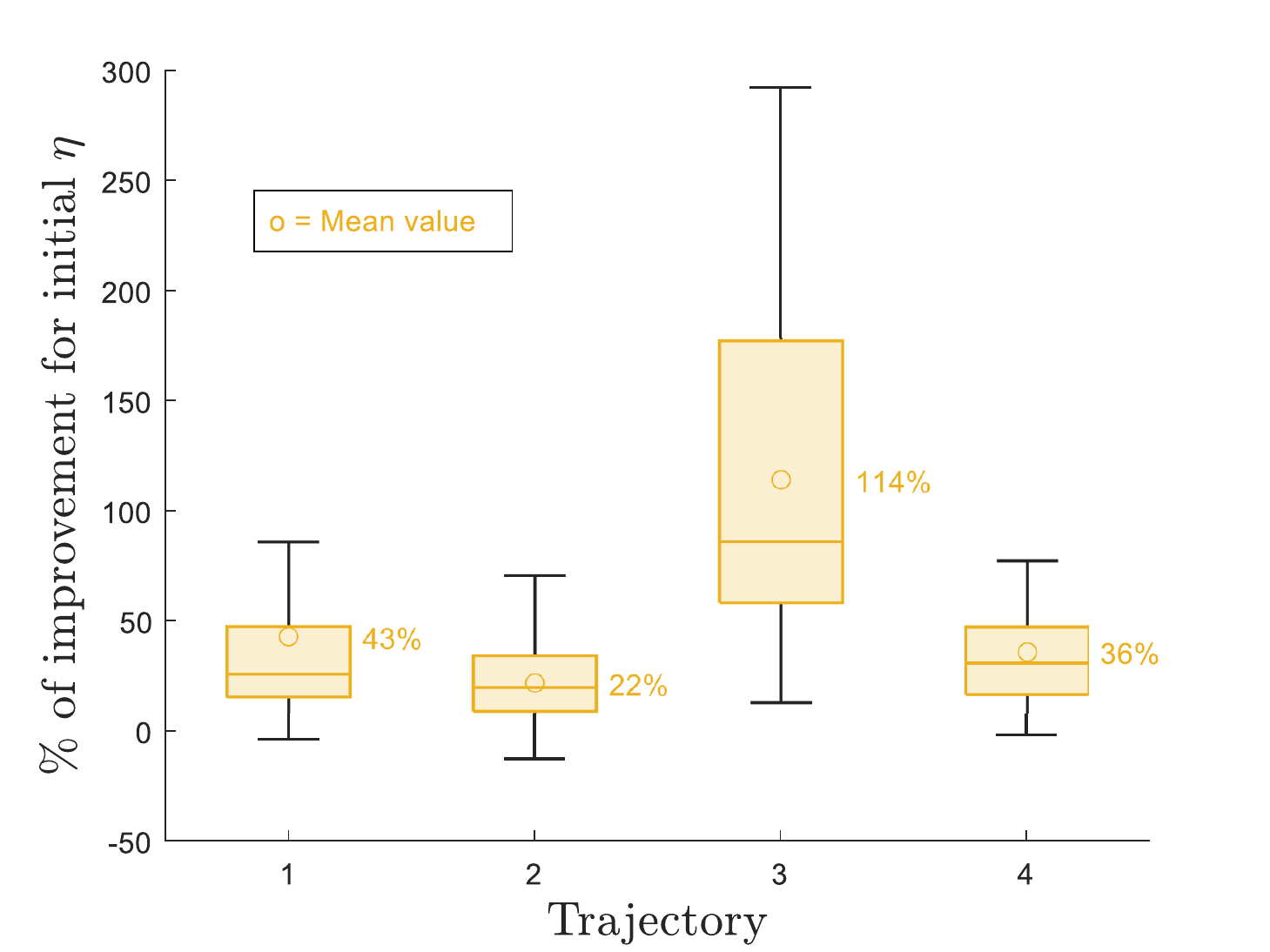}
    \caption{Percentages of optimization for~$\eta$ in the starting pose of each trajectory comparing the optimized and non-optimized case for each of the hundred repetitions. The circle~$\circ$ highlights the mean percentage of the hundred repetitions.}
\label{FIG::percent_eta_init}
\end{minipage}
\hfill
\begin{minipage}[t]{.49\textwidth}
 	\centering
    \includegraphics[scale=0.54]{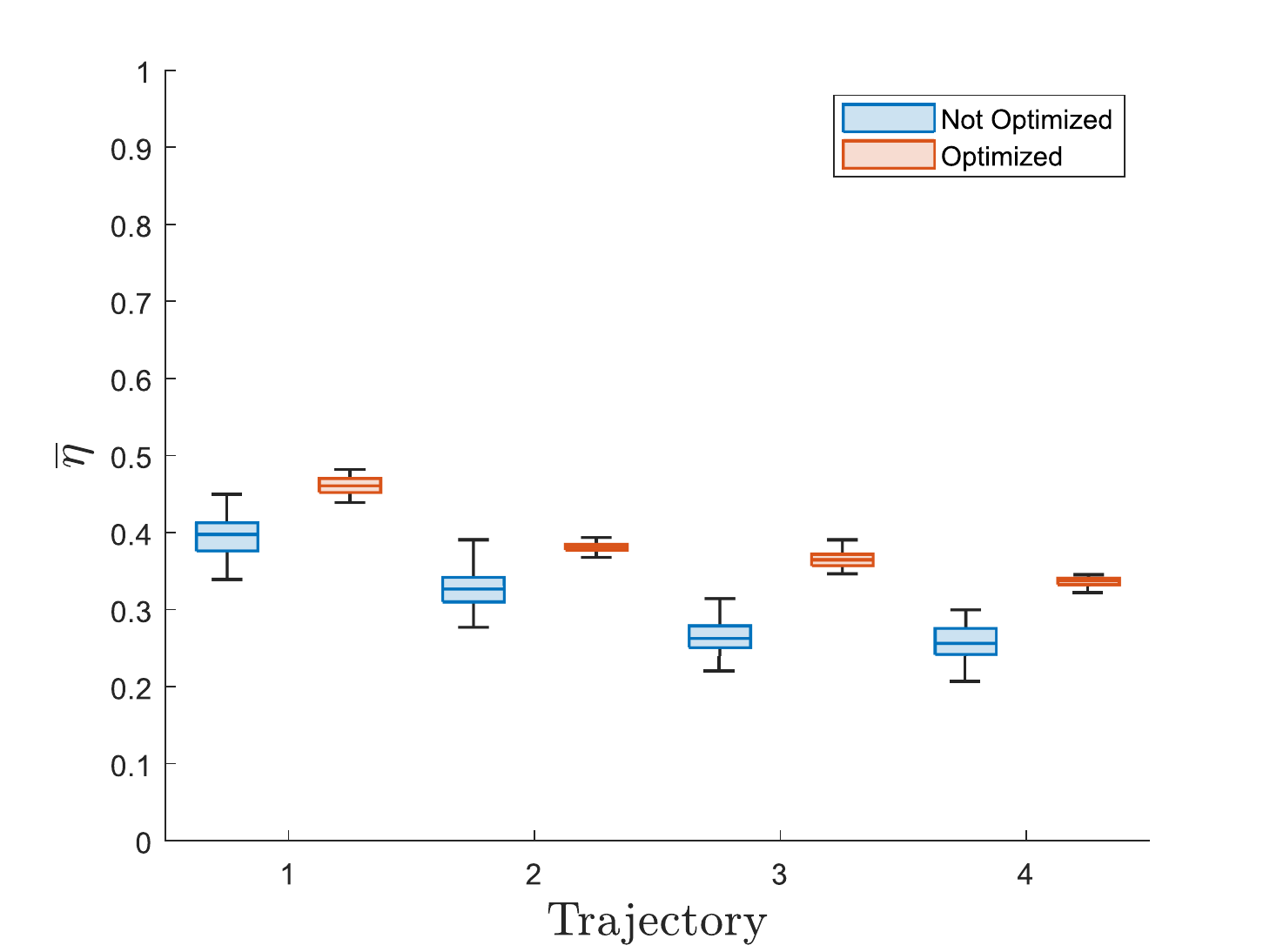}
    \caption{Mean values~$\overline{\eta}$ of~$\eta$ reached along each trajectory (1~to~4, Fig.~\ref{FIG::robot_and_trajectories}) for each one of the hundred repetitions, without (blue) and with (red) optimization of dexterity and RTR}
    \label{FIG::mean_sigma}
\end{minipage}%
\hfill
\begin{minipage}[t]{.49\textwidth}
	\centering
    \includegraphics[scale=0.55]{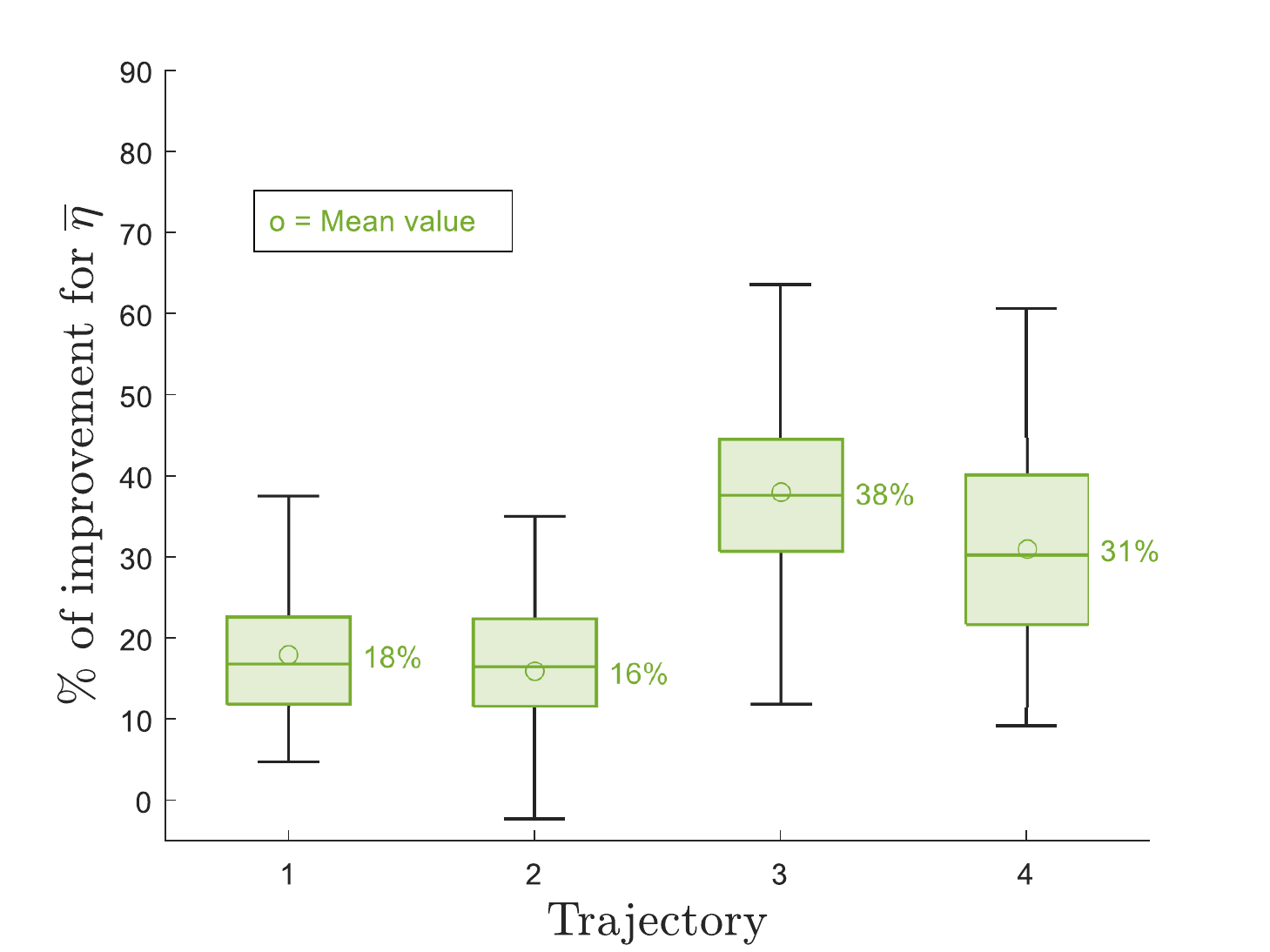}
    \caption{Percentages of optimization for mean value~$\overline{\eta}$ of~$\eta$ on each trajectory comparing the optimized and non-optimized case for each of the hundred repetitions. The circle~$\circ$ highlights the mean percentage of the hundred repetitions.}
\label{FIG::percent_eta}
\end{minipage}
\end{figure}
\begin{figure}[!p]
\centering
\begin{minipage}{.99\textwidth}
\centering
\begin{subfigure}{.49\textwidth}
   	\centering
    \includegraphics[scale=0.6]{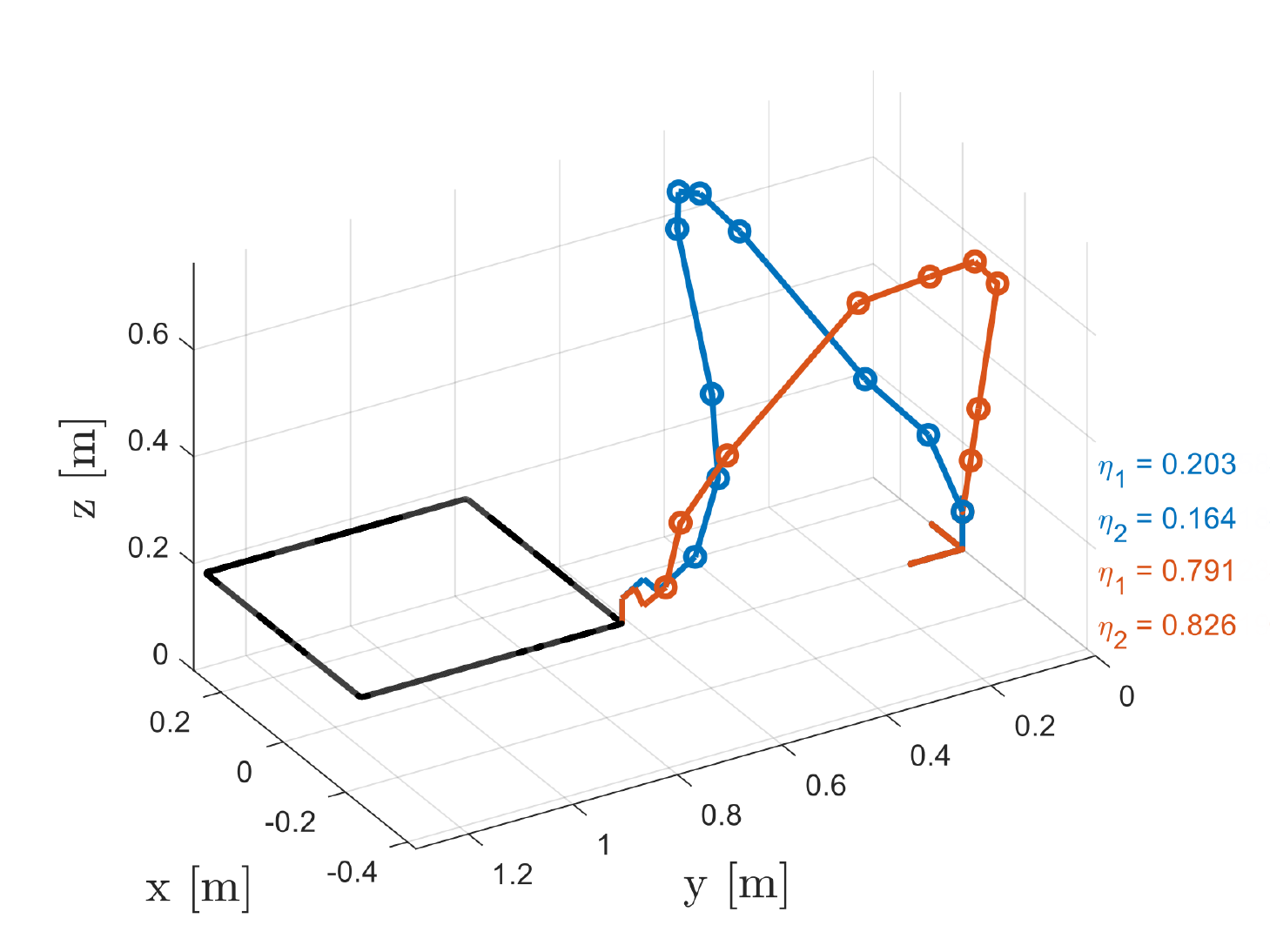}
    \caption{First trajectory}
    \label{FIG::init_config1}
\end{subfigure}%
\hfill
\begin{subfigure}{.49\textwidth}
 	\centering
    \includegraphics[scale=0.6]{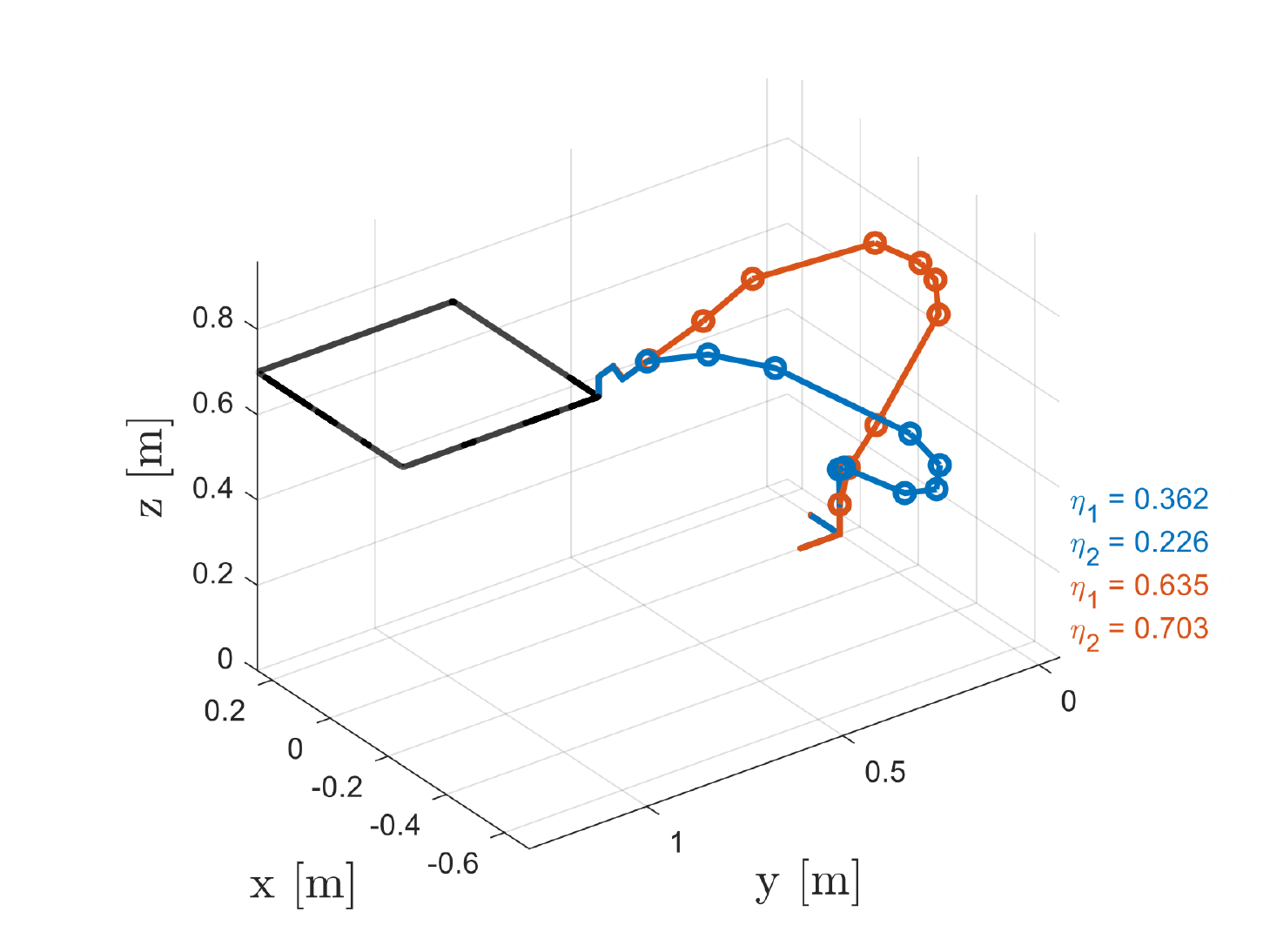}
    \caption{Second trajectory}
    \label{FIG::init_config2}
\end{subfigure}%
\hfill
\begin{subfigure}{.49\textwidth}
   	\centering
    \includegraphics[scale=0.6]{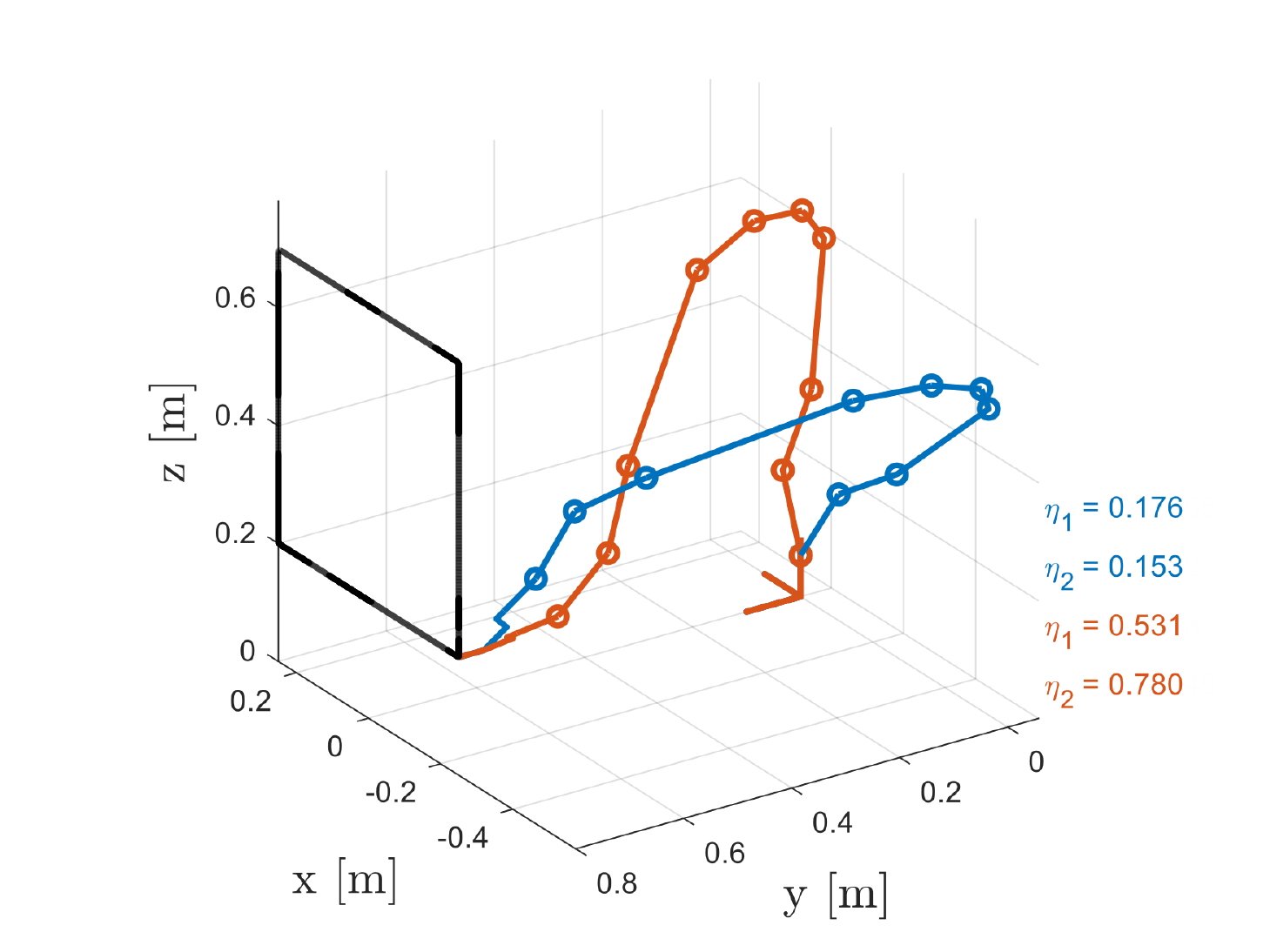}
    \caption{Third trajectory}
    \label{FIG::init_config3}
\end{subfigure}%
\hfill
\begin{subfigure}{.49\textwidth}
 	\centering
    \includegraphics[scale=0.6]{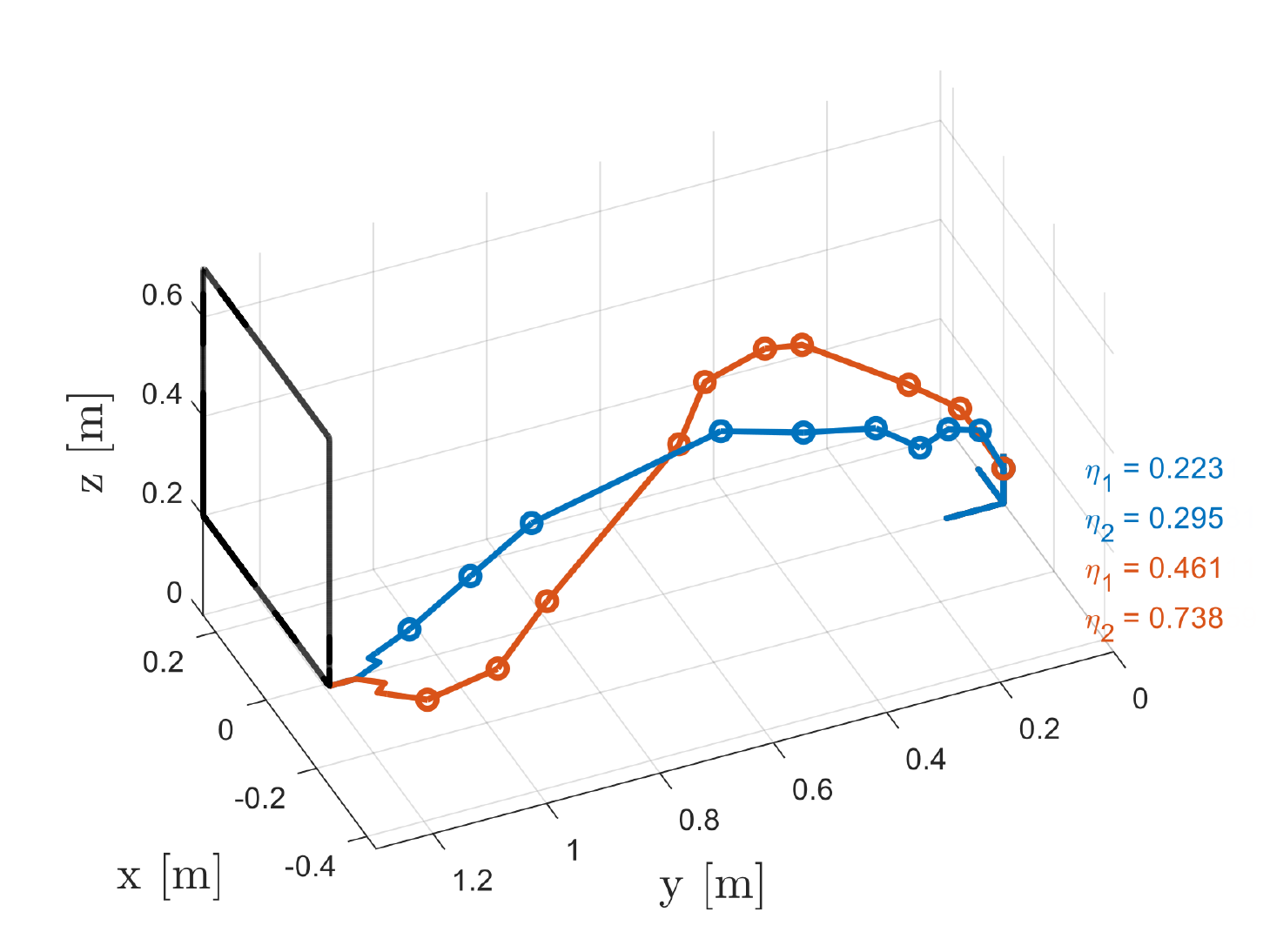}
    \caption{Fourth trajectory}
    \label{FIG::init_config4}
\end{subfigure}%
\caption{Robot in starting configuration on each trajectory for least value of~$\eta$ in case of no optimization (blue) and highest value of~$\eta$ when optimization is active (red). Values of~$\eta_1$ and~$\eta_2$ in both cases are shown on the right of each sub-figure.}
\label{FIG::init_config}
\end{minipage}%
\hfill
\begin{minipage}{.99\textwidth}
\centering
\begin{subfigure}[b]{.49\textwidth}
    \centering
    \includegraphics[scale=0.55]{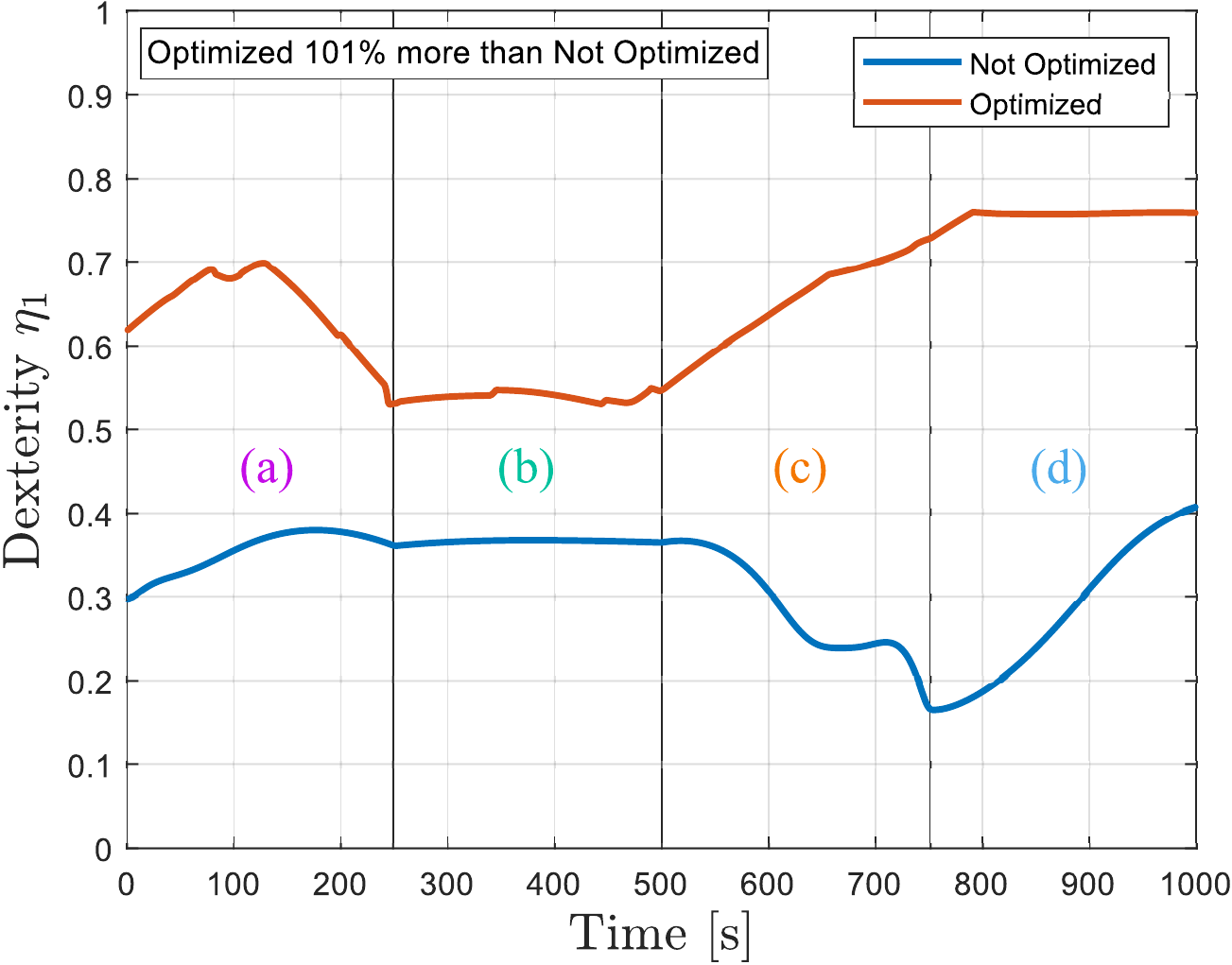}
    \caption{Dexterity metric~$\eta_1$}
\end{subfigure}%
\hfill
\begin{subfigure}[b]{.49\textwidth}
    \centering
    \includegraphics[scale=0.55]{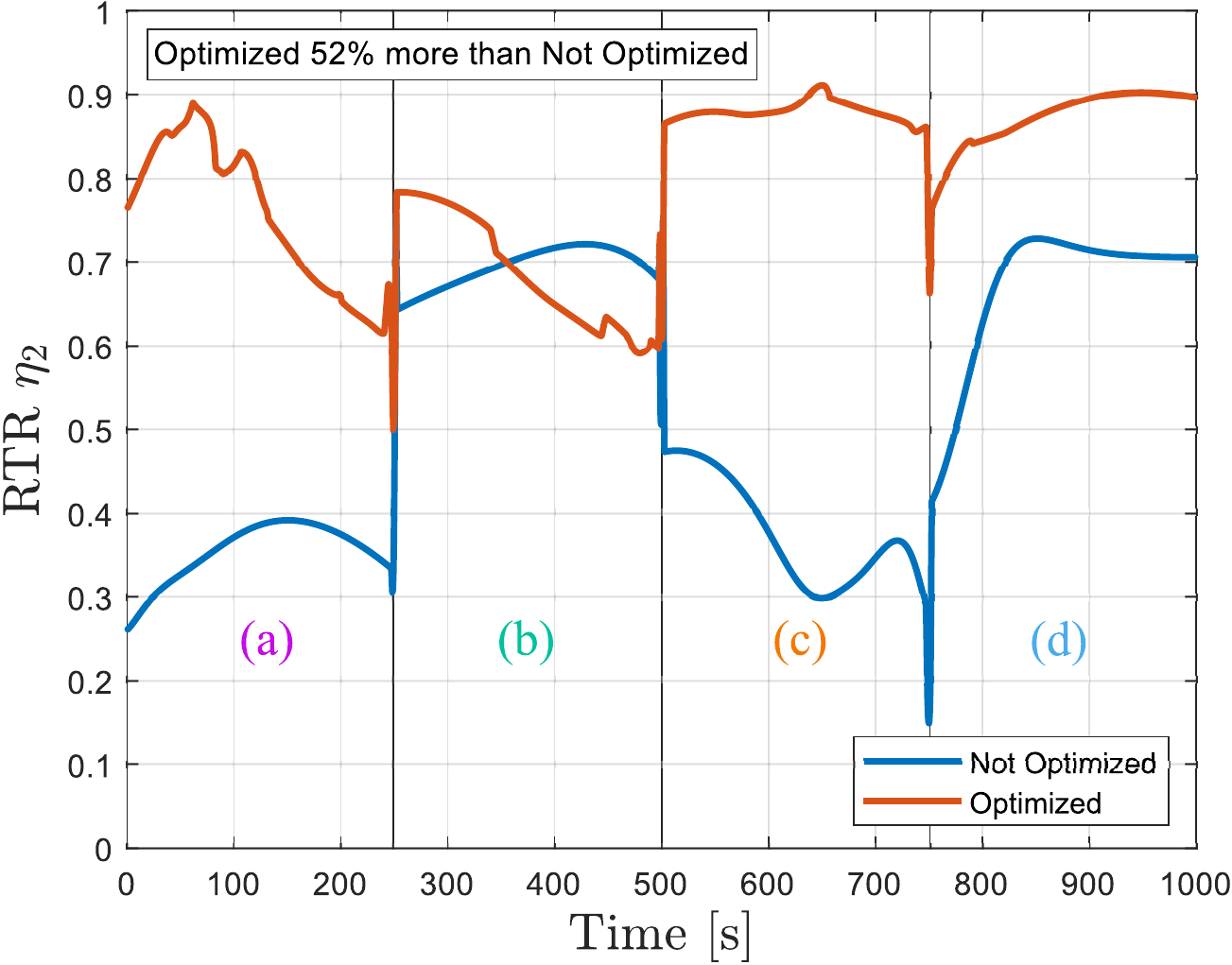}
    \caption{RTR metric~$\eta_2$}
\end{subfigure}%
\caption{Graph of~$\eta_1$ and~$\eta_2$ while following the first trajectory for least value of~$\overline{\eta}$ in case of no optimization (blue) and highest value of~$\overline{\eta}$ when optimization is active (red). Each sector is labeled~(a),~(b),~(c) and~(d) to match the corresponding part of the trajectory.}
\label{FIG::metrics_on_trajectory1}
\end{minipage}%
\end{figure}
\begin{figure}[!t]
\centering
\begin{minipage}{.49\textwidth}
	\centering
    \includegraphics[scale=0.52]{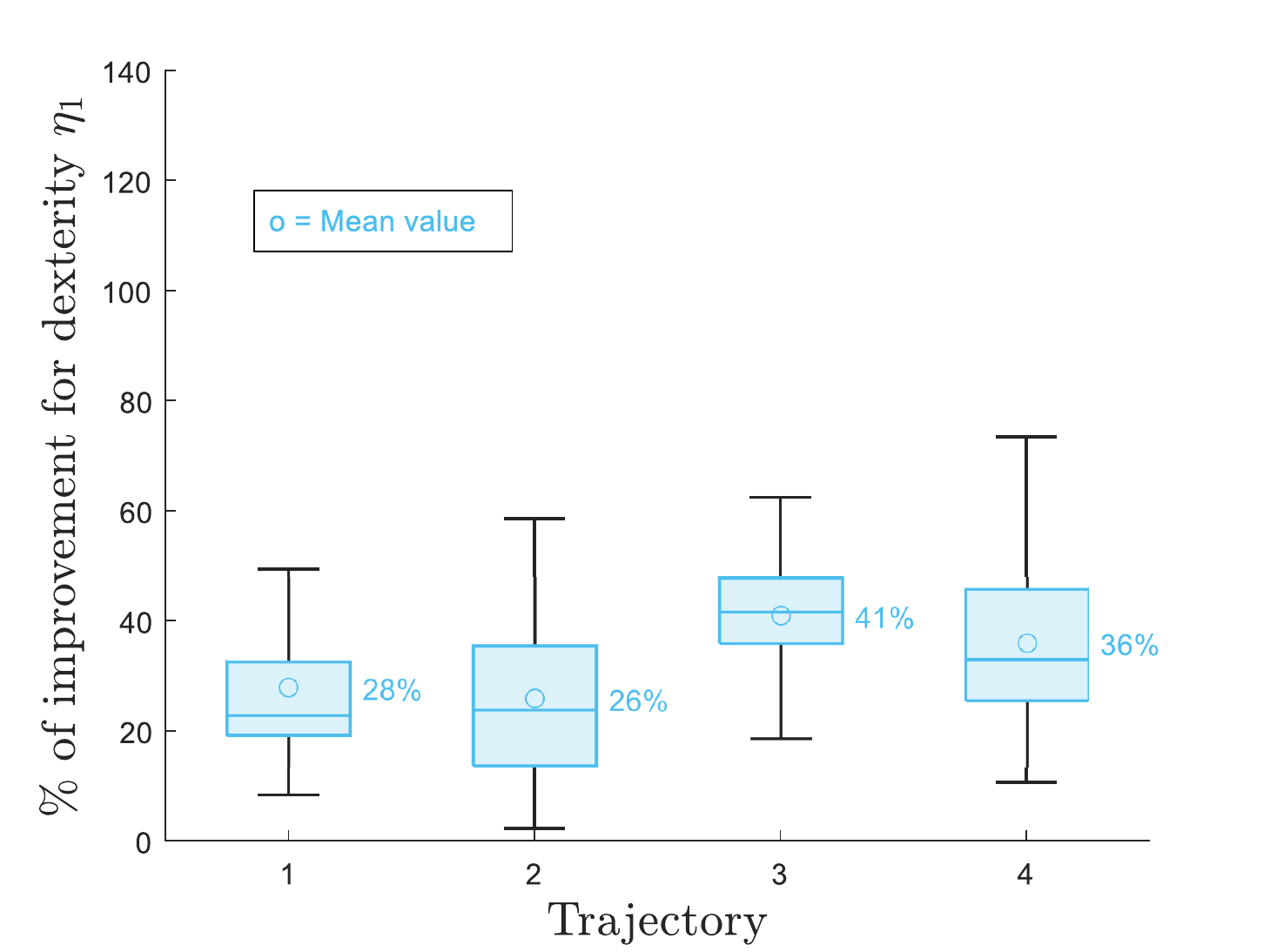}
    \caption{Percentages of optimization for mean value~$\overline{\eta}_1$ of~$\eta_1$ on each trajectory comparing the optimized and non-optimized case for each of the hundred repetitions. The circle~$\circ$ highlights the mean percentage of the hundred repetitions.}
\label{FIG::percent_dxt}
\end{minipage}
\hfill
\begin{minipage}{.49\textwidth}
	\centering
    \includegraphics[scale=0.52]{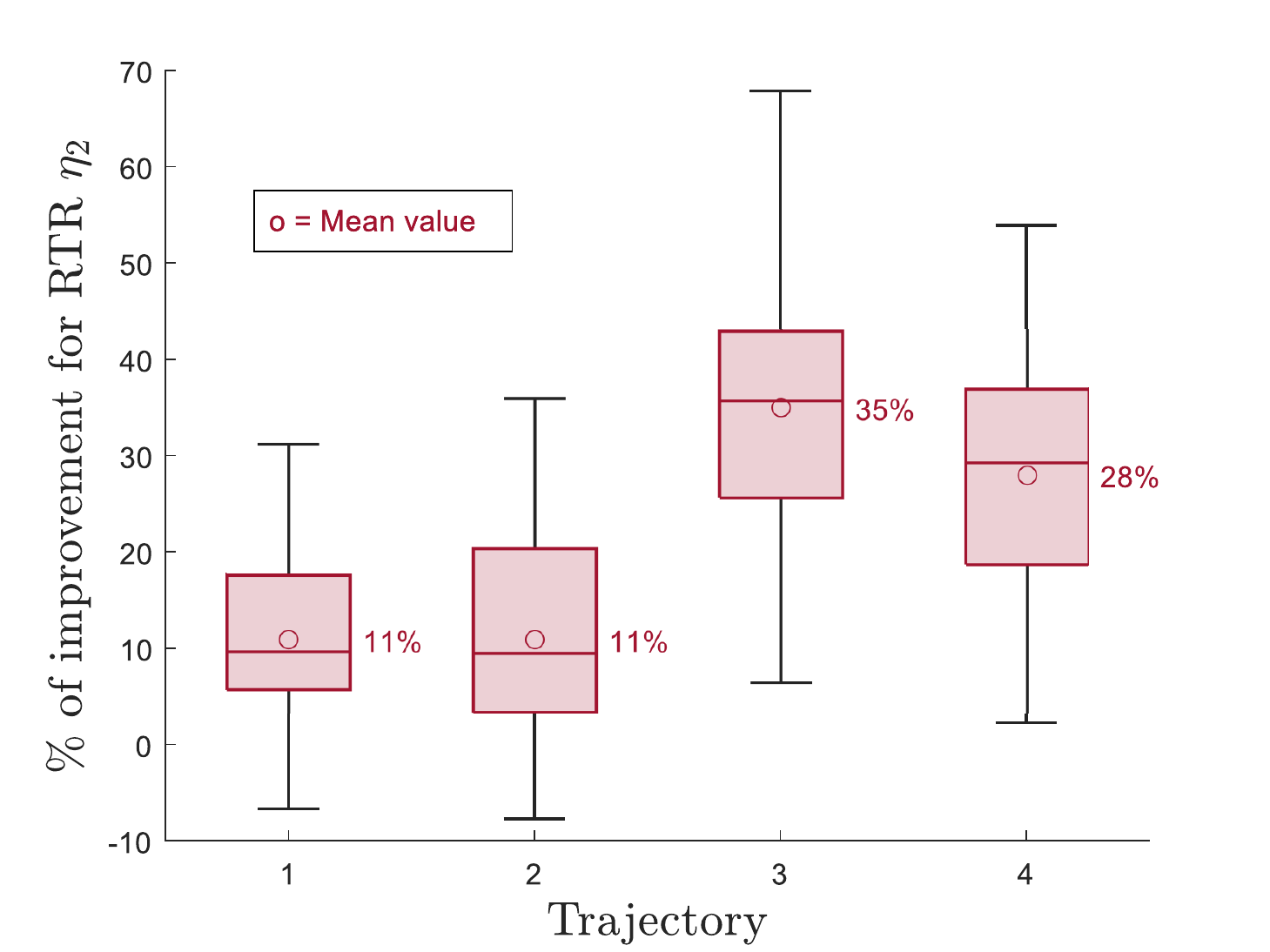}
    \caption{Percentages of optimization for mean value~$\overline{\eta}_2$ of~$\eta_2$ on each trajectory comparing the optimized and non-optimized case for each of the hundred repetitions. The circle~$\circ$ highlights the mean percentage of the hundred repetitions.}
\label{FIG::percent_rtr}
\end{minipage}
\end{figure}

The tests are developed as follows.
The robot is started from a random configuration and reaches the starting point on one trajectory.
From here, it tracks the entire trajectory and collects dexterity and RTR values at each step.
These actions are repeated hundred times starting from different random robot configurations to obtain a general pool of results.
Then, the same process is repeated activating the dexterity and RTR tasks, both when approaching the starting point and following the trajectory.
The robot is initialized using the same random configurations used in the tests without optimization.
Moreover, when the optimization tasks are active, a monitoring is added while reaching the starting point to check if the control algorithm is still optimizing the robot configuration even though the end-effector frame has already reached the desired pose.
This methodology is applied to each trajectory.
A summary of the methodology is presented in Algorithm~\ref{ALG::method}.

Figures~\ref{FIG::init_sigma} and~\ref{FIG::mean_sigma} give a general overview of the results.
Figure~\ref{FIG::init_sigma} collects the values of~$\eta$ of the robot on the starting poses with and without optimization tasks.
For each trajectory~1 to~4, two box plots are shown containing the results obtained by repeating the process without (blue) and with (red) optimization a hundred times.
When the dexterity and RTR tasks are activated, the variance of~$\eta$ is smaller and the minimum and maximum values are high.
This means that the optimization tasks always help reaching configurations with high values of~$\eta$.
On the contrary, when the dexterity and RTR tasks are deactivated, the variance of~$\eta$ on the starting poses is bigger.
Since the optimization tasks are disabled, the TPIK algorithm runs the robot straightly to the desired pose without performing any optimization on the robot configuration and~$\eta$ can reach higher or lower values.
Furthermore, as described in the testing algorithm, the action~$\mathscr{A}_3$ is allowed to run until the metrics local maxima are reached.
This implies that moving the robot to the starting pose with active dexterity and RTR tasks requires, on average, 80\% more time than without these tasks; even though the end-effector reaches the desired pose in the same time in both scenarii.
In fact, the average time for running the TPIK algorithm at each step is~$698\;\mu\text{s}$ when the optimization tasks are disabled and~$744\;\mu\text{s}$ when the optimization is active.
Figure~\ref{FIG::percent_eta_init} presents the improvement of~$\eta$ at the starting pose of each trajectory, with and without the optimization tasks for each of the hundred repetitions.
The improvement of~$\eta$ are almost always high, over 250\% in some cases.
However, few cases show a negative percentage.
This is due to the control algorithm converging to a local maxima when optimizing the metrics.
In fact, the optimized control algorithm reaches a local maxima in these few cases while the non-optimized one moves the robot in a configuration that escaped the local maxima and had higher kinetostatic performance unintentionally.
This behavior happens a few times, which justifies the need to run the algorithm several times to obtain the best robot performance.

Figure~\ref{FIG::mean_sigma} collects the mean values~$\overline{\eta}$ of~$\eta$ reached on each trajectory.
Again, for each trajectory~1 to~4, there are two box plots containing the results obtained by repeating a hundred times the process without (blue) and with (red) optimization.
The optimization tasks lead to a smaller variance for~$\overline{\eta}$ compared to the non-optimized results.
However, the minimum values of~$\overline{\eta}$ in the non-optimized cases are higher than the minimum~$\eta$ obtained on the non-optimized starting poses.
This means that the RP-120 can maintain good kinetostatic performance on the trajectories even if the optimization tasks are disabled and the starting~$\eta$ is low.
Moreover, the first trajectory shows the best performance, followed by the second, the third and the fourth.
So, this design has better kinetostatic performance when the trajectory is horizontal and nearer to the base.
Figure~\ref{FIG::percent_eta} presents the improvement of~$\eta$ on each trajectory, with and without the optimization tasks for each of the hundred repetitions.
Again, it can be noticed that in a few cases the percentage is negative due local maxima issue.

Figure~\ref{FIG::init_config} shows the robot on the starting pose of each trajectory for the minimum value of~$\eta$ in case of no optimization and the maximum~$\eta$ when optimization tasks were active.
The values~$\eta_1$ and~$\eta_2$ for the optimized and not optimized configurations are shown next to the robots in each figure.
In the configurations assumed by the optimized robots, the~$\vec{x}$ pose difference of one NB-module center and the next is lower than in the non-optimized cases.
This leads to a smaller angle between the joint velocity and torque vectors and increases the RTR value.
Moreover, the vectors from the joint frames to the end-effector frame are different in length and orientation, giving distinct contributions to the kinematic Jacobian matrix and increasing the dexterity value.

Figure~\ref{FIG::metrics_on_trajectory1} shows the robot dexterity and RTR profiles along the first trajectory for the minimum value of~$\overline{\eta}$ in case of no optimization and the maximum~$\overline{\eta}$ when optimization tasks are active.
The dexterity and RTR graphs are divided in the four trajectory sectors~(a),~(b),~(c) and~(d).
The curves are higher when their tasks are active.
The graphs also show the percentage of optimization for each curve.
The activation of the dexterity task can improve the performance by a factor $2$.
It can also be noticed how the RTR curve has discontinuities in correspondence to the trajectory corners since its value is directly affected by the orientation of the velocity and force vectors applied to the ending tool.
Here, only the graphs for the first trajectory are shown since all the other trajectories showed similar behaviors.
However, a video of the robot simulation on each trajectory with dexterity and RTR graphs for minimum value of~$\overline{\eta}$ without optimization and maximum~$\overline{\eta}$ with optimization can be seen at this link\footnote{Link to simulation video: \url{https://nimbl-bot.com/video-nb-120_1/}}.
Comparing all the dexterity and RTR results on each trajectory with and without optimization repeated a hundred times shows that the optimization tasks averagely increase the dexterity value of~33\% and the RTR of~22\%.
Figures~\ref{FIG::percent_dxt} and~\ref{FIG::percent_rtr} show the improvement of~$\overline{\eta}_1$ and~$\overline{\eta}_2$ respectively, with and without the optimization tasks on all trajectories and for a hundred repetitions.
These percentages demonstrate how the use of optimization tasks generally improves the robot performance.
In few cases, the negative percentage issue is met.
However, this issue appears only for the RTR.

A final consideration that can be pointed out is the relation between the RTR value and the velocity and force vectors orientation.
Taking into account the first and second trajectories, the RTR values are~24\% higher when the robot moves along~$\vec{x}$, (b) and~(d) sectors, than along~$\vec{y}$, (a) and~(c) sectors, in case of no optimization.
When the RTR task is active, the percentage reduces to~16\% because the task helps optimizing the robot configuration.
Then, considering the third and fourth trajectories, the RTR values are~21\% higher when the robot moves along~$\vec{x}$, (b) and~(d) sectors, than along~$\vec{z}$, (a) and~(c) sectors.
When the RTR task is active, the percentage becomes~3\%.
So, the robot has higher performance for the RTR when the end-effector does a tangential horizontal movement than radial and vertical movements.
A similar behavior can not be noticed in the dexterity because it is not affected by the magnitude or orientation of the velocity and force vectors applied to the end-effector.

\section{Conclusions and Future Work}
\label{SEC::conc}

This paper presented a new design for redundant robots formed of a series of closed kinematic chain mechanisms called NB-modules.
The geometric and kinematic models of the NB-module are extracted from the previous work and presented.
Then, a kinematic control algorithm called Task Priority Inverse Kinematic~(TPIK) is used to kinematically control the robot performing machining tasks.
This kinematic control algorithm is suitable for the RP-120 and exploits its kinematic redundancy to solve simultaneous tasks.
Two new tasks are introduced to improve the robot kinetostatic performance.
One is based on the dexterity and the other on the robot transmission ratio~(RTR).
The robot tracks a series of trajectories, both activating and deactivating the optimization tasks.
The major limitation of this algorithm is that it is attracted by local maxima.
So, the process needs to be run several times to come up with the best joint trajectory planning for the desired set of tasks.
When the optimization tasks are active, the results clearly show an improvement in the robot performance, both dexterity and RTR, without affecting the time consumption.
In fact, the average time consumed by the algorithm at each step with or without the optimization tasks is almost equal, the difference is less than~$50\;\mu\text{s}$.
To rate the improvement given by the optimization tasks, a linear combination of dexterity and RTR is used, i.e.~$\eta$.
When the robot reaches the trajectory starting pose with active optimization task,~$\eta$ is averagely~54\% higher than the non-optimized case.
Along each trajectory, the mean value~$\overline{\eta}$ is averagely~26\% higher in the optimized case compared to the non-optimized one.
Moreover, it can be noticed that the kinetostatic performance are affected by the trajectory placement and by the velocity and force vectors orientation.
However, this paper uses only four squared trajectories with constant velocities and forces to test the robot performance.
So, it is not possible to identify the best placement and orientation for the workpiece, but this is a starting point for future analysis.
Then, there is no comparison with the performance of other existing manipulators on the same tasks.
This will be done later on to study the potential of the RP-120 with respect to its counterparts.
Finally, the prototype shown in Section~\ref{SEC::nimb_rob} will be tested to analyze its performance in real life.


\begin{acknowledgment}
This work was supported by the ANRT (Association Nationale de la Recherche et de la Technologie) grant CIFRE n$^{\circ}$ 2020/1051 and Nimbl'Bot (\url{https://nimbl-bot.com/})
\end{acknowledgment}



%
% ...



\listoffigures

\listoftables


\end{document}